\RequirePackage{fix-cm}
\documentclass[twocolumn]{svjour3}
\smartqed
\usepackage{times}
\usepackage{graphicx}
\usepackage{arydshln}
\usepackage{amsmath, amssymb, bm}
\usepackage{booktabs, multirow, threeparttable}
\usepackage{caption, subcaption}
\usepackage[dvipsnames,table]{xcolor}
\usepackage{colortbl}
\usepackage{makecell}
\usepackage{siunitx}

\usepackage{cite}
\usepackage{calc}
\usepackage{xspace}
\usepackage{orcidlink}
\usepackage{wasysym, utfsym, marvosym}
\usepackage{stfloats}
\usepackage{verbatim}
\usepackage{float} 

\usepackage{hyperref}
\hypersetup{
	colorlinks=true,
	linkcolor=red,
	filecolor=blue,      
	urlcolor=red,
	citecolor=blue,
}

\usepackage{array}% 用于 >{...} 语法 
\newcounter{rowno}%创建一个行号计数器
\newcommand{\condcolor}{\ifodd\value{rowno}\else\cellcolor{gray!10}\fi}
\newcolumntype{C}{>{\condcolor}c}

\definecolor{DeepOliveGreen}{RGB}{50,90,0}

\makeatletter
\DeclareRobustCommand\onedot{\futurelet\@let@token\@onedot}
\def\@onedot{\ifx\@let@token.\else.\null\fi\xspace}
\def\eg{\emph{e.g}\onedot} 
\def\ie{\emph{i.e}\onedot}

\makeatother
\begin{document}
\sloppy

%\title{Image Quality Assessment for Machines: From Databases to Benchmarks and Beyond}
\title{Image Quality Assessment for Machines: Paradigm, Large-scale Database, and Models}

\author{Xiaoqi Wang \and 
	Yun Zhang \textsuperscript{\Letter} \and 
    Weisi Lin %\and 
	% Christian Micheloni
}  
%\author{Masato Tamura\orcidlink{0000-0003-1029-5271}}

%\institute{Masato Tamura \at
%              Big Data Analytics Solutions Lab, R\&D, \\
%              Hitachi America, Ltd., \\
%              2535 Augustine Dr, 3rd Floor, Santa Clara, 95054, California, United States. \\
%              \email{masato.tamura@ieee.org}
\institute{
	Xiaoqi Wang, Yun Zhang \at
	School of Electronics and Communication Engineering, Sun Yat-sen University,
	Shenzhen, 518107, China. \\
	\email{wangxq79@mail2.sysu.edu.cn; zhangyun2@mail.sysu.edu.cn}
	\and
	Weisi Lin \at
	College of Computing and Data Science, Nanyang Technological University,
	50 Nanyang Avenue, 639798, Singapore. \\
	\email{wslin@ntu.edu.sg}
}
%\date{}%Received: date / Accepted: date}
\maketitle
 
\begin{abstract} 

Machine vision systems (MVS) are intrinsically vulnerable to performance degradation under adverse visual conditions. To address this, we propose a machine-centric image quality assessment (MIQA) framework that quantifies the impact of image degradations on MVS performance. We establish an MIQA paradigm encompassing the end-to-end assessment workflow. To support this, we construct a machine-centric image quality database (MIQD-2.5M), comprising 2.5 million samples that capture distinctive degradation responses in both consistency and accuracy metrics, spanning 75 vision models, 250 degradation types, and three representative vision tasks. We further propose a region-aware MIQA~(RA-MIQA) model to evaluate MVS visual quality through fine-grained spatial degradation analysis. Extensive experiments benchmark the proposed RA-MIQA against seven human visual system (HVS)-based IQA metrics and five retrained classical backbones. Results demonstrate RA-MIQA’s superior performance in multiple dimensions, \eg, achieving SRCC gains of 13.56\% on consistency and 13.37\% on accuracy for image classification, while also revealing task-specific degradation sensitivities. Critically, HVS-based metrics prove inadequate for MVS quality prediction, while even specialized MIQA models struggle with background degradations, accuracy-oriented estimation, and subtle distortions. This study can advance MVS reliability and establish foundations for machine-centric image processing and optimization. The model and code are available at:~\url{https://github.com/XiaoqiWang/MIQA}.  
\end{abstract}

\section{Introduction}
The rapid advancement of intelligent algorithms and computational power has catalyzed a fundamental transformation in computer vision, enabling machines to achieve superhuman performance in machine tasks. This progress has driven the wide deployment of automated visual systems across numerous applications, including surveillance~\cite{delussu2024synthetic}, autonomous vehicles~\cite{autonomous_driving_survey}, and edge intelligence~\cite{edgenext}.

\begin{figure}[t!] \centering 
	\includegraphics[width=0.49\textwidth]{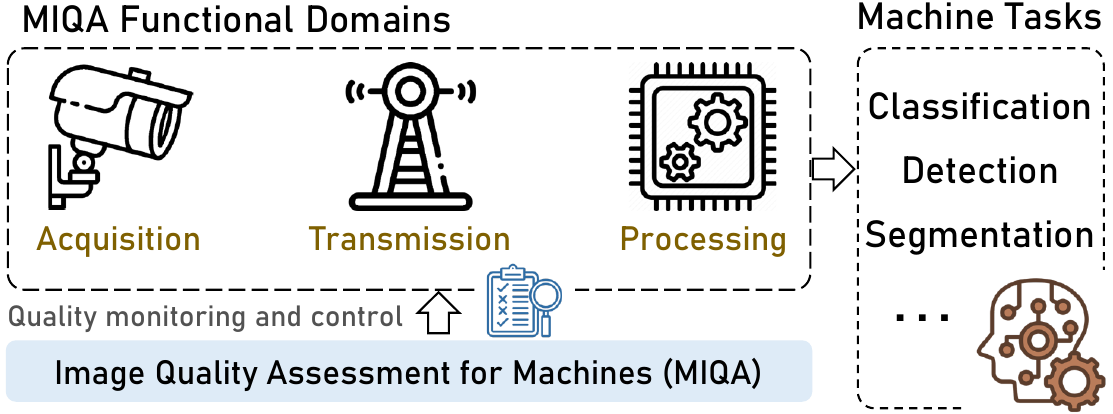}   
	\caption{The applications of MIQA in MVS: quality monitoring and control across image acquisition, transmission, and processing stages to optimize downstream vision tasks.} \vspace{-1em}
	 \label{fig:func}
\end{figure} 

However, the effectiveness of these systems is critically undermined by image degradations commonly encountered in real-world environments~\cite{2018generalisation, tpami_effects_CNNs}. These degradations originate from diverse sources, including acquisition conditions~(\eg, illumination changes, motion blur, and adverse weather), transmission artifacts~(\eg, packet loss and signal corruption), and processing-induced distortions~(\eg, compression and resolution downscaling). Such degradations can significantly impair model predictions, potentially resulting in economic losses and, more critically, posing life-threatening risks in safety-critical applications~(\eg, autonomous driving~\cite{enhancement_auto_driv} and medical imaging~\cite{radiography_effect}). A growing body of research has advanced machine-oriented image processing algorithms that prioritize downstream task performance, including joint denoising and semantic learning frameworks~\cite{liu2018_ijcai_denoising}, dual-branch dehazing models for object detection~\cite{Li2023_Dehazing}, task-preserving compression preprocessing~\cite{preprocess_compression}, and diffusion-based compressed image restoration with mixture-of-experts prompting~\cite{Moe_diffir}. However, these processing approaches lack dedicated machine-centric quality metrics at the image level, hindering the analysis of degradation impacts on machine performance and the corresponding optimization of processing strategies.

In this context, image quality assessment (IQA) emerges as a crucial mechanism for monitoring and controlling visual degradations across acquisition, transmission, and processing stages, playing a key role in safeguarding downstream task performance for machine vision systems (MVS) (as shown in Figure~\ref{fig:func}). While IQA for the human visual system~(HVS) is a long-standing topic, it prioritizes human perceptual fidelity~(\eg, chromatic accuracy) with evaluation criteria grounded in human visual experience~\cite{ssim, Kalpana_tip2010}. In contrast, MVS exhibits distinct degradation sensitivities~\cite{2022jin_jrd}, including a region-specific vulnerability in areas crucial for machine analysis~(\eg, edges and keypoints) and susceptibility to subtle perturbations that are imperceptible to humans yet detrimental to feature detection~\cite{2019CVPR_Transferability}. This fundamental disparity between human and machine perception reveals the inherent limitations of HVS-based IQA methods for MVS, thereby catalyzing efforts toward developing machine-based IQA~(MIQA) designed for evaluating visual degradation impacts on MVS~\cite{venkataramanan2022assessing, ob_ImageQA}. 

The development of effective MIQA methods faces key challenges in establishing quality metrics that correlate with machine performance, accounting for task-specific sensitivities, addressing the diversity of potential degradations, and ensuring generalization across model architectures. Addressing these challenges would not only deepen our understanding of machine perception mechanisms but also enable critical applications in controlled-quality image acquisition, processing optimization, and enhanced reliability of MVS in adverse conditions. Building on this foundation, we made the following contributions. %Based on this foundation, our key contributions advance the field through: 
\begin{enumerate}
	\item[(i)]We establish a methodological framework for the MIQA task, which comprises key components: model spaces, data spaces, and label spaces with consistency and accuracy metrics, while formalizing protocols for MIQA model validation.
	\item[(ii)] We develop a large-scale machine-centric image quality database, MIQD-2.5M, comprising 2.5 million samples across 10 degradation types, 5 intensity levels, and 3 regional perturbation patterns. We record the degradation responses of 75 vision models across three tasks, then construct and validate machine-oriented quality labels.  
	
	\item[(iii)]We propose a degradation \textbf{r}egion-\textbf{a}ware MIQA~(called RA-MIQA) model and benchmark it against existing HVS-based IQA methods and five representative models retrained on MIQD-2.5M, demonstrating superior performance across multiple dimensions, including label types, degradation regions, distortion categories, intensity levels, and cross-task generalization.
\end{enumerate}

The remainder of this paper is organized as follows: Section~\ref{sec:related_work} reviews related work on machine vision in degraded conditions. Section~\ref{sec:formulations} formally defines the MIQA task. Section~\ref{sec:database} describes the construction, statistical analysis, and label validation of the MIQD-2.5M. Section~\ref{sec:method} presents the proposed MIQA model. Section~\ref{sec:benchmark} benchmarks the proposed method against HVS-based IQA and MIQA models, and discusses the limitations of MIQA methods. Finally, Section~\ref{sec:conc} concludes the paper.

\section{Related Work}
\label{sec:related_work}
Maintaining the stability of machine performance under degraded visual inputs involves two dimensions: assessing model robustness against various visual distortions and evaluating whether input quality meets the requirements of downstream models. This section reviews the relevant literature from both perspectives.
\subsection{Model Robustness Evaluation}

To cope with image degradations encountered in real-world scenarios, recent studies have investigated model robustness against image corruptions~\cite{hendrycks2019robustness, tpami_effects_CNNs, 2018generalisation, radiography_effect,kamann2021benchmarking,liu2024benchmarking}. Hendrycks~\textit{et al}.~\cite{hendrycks2019robustness} introduced IMAGENET-C and IMAGENET-P, revealing limited robustness gains from architectural advancements alone and demonstrating the benefits of techniques like histogram equalization, larger models, and adversarial training. Pei~\textit{et al}.~\cite{tpami_effects_CNNs} investigated the impact of various image degradations on CNN-based classification models, observing significant performance drops when training and test degradations differ, and the limited efficacy of traditional restoration methods. Geirhos~\textit{et al}.~\cite{2018generalisation} compared human and DNN object recognition under diverse degradations, highlighting superior human robustness and limited DNN generalization. Sabottke \textit{et al}.~\cite{radiography_effect} explored the influence of image resolution on CNN-based medical image diagnosis, identifying optimal resolutions for specific tasks. Kamann \textit{et al}.~\cite{kamann2021benchmarking} showed that segmentation robustness depends on corruption type and can be improved by architectural elements such as atrous convolutions and skip connections. Schwartz \textit{et al}.~\cite{liu2024benchmarking} exposed widespread vulnerabilities in 12 object detectors under 16 sensor-based corruptions. Although these studies focus on model-level architectural robustness, assessing image-level quality is equally essential to maintaining operational reliability under degradation. Therefore, effective image quality evaluation is indispensable for mitigating task performance degradation in downstream machine tasks.
\subsection{HVS-based IQA Methods}
HVS-based IQA constitutes a long-standing research paradigm that has evolved along two primary directions. Full-reference (FR) methods, such as PSNR and SSIM~\cite{ssim}, assess pixel-level fidelity, whereas perceptual feature-based metrics like LPIPS~\cite{lpips} and DISTS~\cite{dists} estimate perceptual similarity in learned representation spaces. Concurrently, no-reference (NR) methods have emerged, including NIMA~\cite{nima}, which predicts human opinion score distributions via squared Earth Mover's Distance loss, as well as more recent deep architectures such as the CNN-based HyperIQA~\cite{hyperiqa} and Transformer-based MANIQA~\cite{maniqa}. However, these methods inherently approximate human perceptual judgments rather than task-specific utility for MVS, failing to capture their unique sensitivity patterns. This misalignment is starkly revealed by adversarial examples, where imperceptible perturbations lead to confident misclassifications, indicating that images rated as high-quality by HVS can still impair machine performance under corruption~\cite{FGSM, attacks_Fezza2019}. Such discrepancies underscore the limitations of HVS-based IQA in machine-centric scenarios and motivate the development of MIQA approaches.

\subsection{Machine-based IQA Methods}
Early work related to MIQA can be traced back to biometric sample quality measures~\cite{biometric_tpami}, where quality refers to any factor that impairs machine performance. In face recognition, evaluation centered on the utility of a facial image based on its ability to maintain recognition accuracy under adverse conditions~\cite{survey_face}. The rise of neural network-based vision systems has expanded MIQA research to a broader range of machine tasks. Venkataramanan et al.~\cite{venkataramanan2022assessing} extracted Log-Gabor features from distorted images and employed support vector regression to predict detection performance in degraded road environments. Beniwal et al.~\cite{ob_ImageQA} predicted object detection performance by estimating the ratio of discrete cosine transform coefficients between original and compressed feature representations. Marie~\cite{segmentation_miqa} assessed the correlation between HVS-based IQA methods and the perceptual performance of semantic segmentation models on compressed images, revealing a weak correlation at the image block level. Dremin~\cite{icpr_compressed} proposed a CNN-based full-reference MIQA model that quantifies the impact of various image and video encoders on detection and recognition performance. Concurrent research by Li \textit{et al}.~\cite{li2025iqa_cvpr}, evaluating over 30K distorted images across seven downstream tasks on 30 models, reveals a substantial discrepancy between human- and machine-oriented IQA.

However, these methods are limited by one or more of the following factors: 1)~Model generalizability: Studies \cite{ob_ImageQA, segmentation_miqa} focus on single architectures (\eg, Faster R-CNN\cite{faster_rcnn} and DeepLabV3\cite{DeepLabV3}), introducing architecture-specific biases that impede generalization. 2)~Distortion region: All these methods~\cite{ob_ImageQA, segmentation_miqa, icpr_compressed, li2025iqa_cvpr} assume uniform distortions, neglecting the disproportionate influence of distortions within task-relevant regions (\eg, ROI versus background). 3)~Task sensitivity: Differences in quality sensitivity across tasks, particularly between coarse-grained tasks (\eg, image classification) and fine-grained tasks (\eg, object detection or instance segmentation), remain underexplored. 4)~Quality label: MIQA labels are based on either consistency (alignment between degraded and original image predictions)~\cite{segmentation_miqa, icpr_compressed, li2025iqa_cvpr} or accuracy (correctness of degraded image predictions against ground truth)~\cite{venkataramanan2022assessing}, yet these metrics represent distinct and equally significant dimensions for MVS.

\section{MIQA Paradigm}
\label{sec:formulations}
\begin{figure}[t]
	\centering  
	\includegraphics[width=0.49\textwidth]{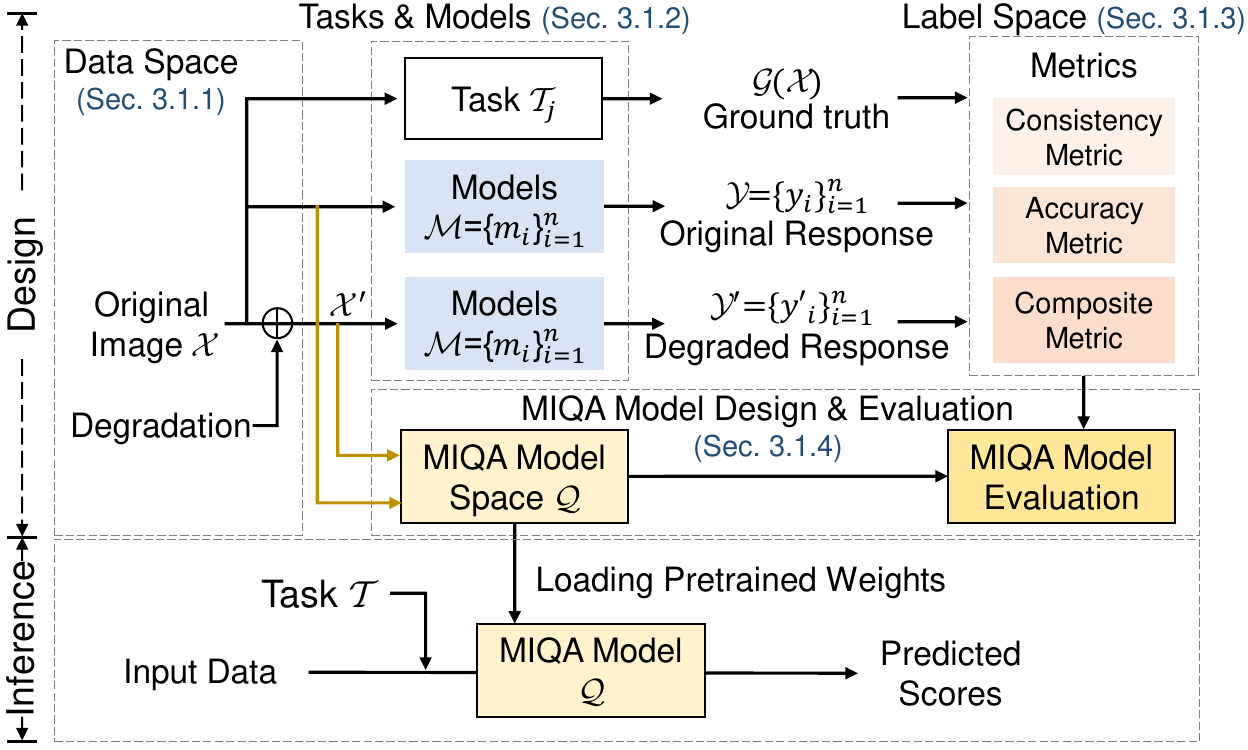}  \vspace{-1.5em}
	\caption{The paradigm of MIQA.} \label{fig:formulation}
	\vspace{-1em}
\end{figure}

As shown in Figure~\ref{fig:formulation}, the MIQA paradigm is based on three interconnected components: Data space (Section~\ref{sec:input_space}) tracking image distortion processes, vision task and model space (Section~\ref{sec:model_task_space}) analyzing model behavior under various conditions, and label space (Section~\ref{sec:label_space}) quantifying quality through consistency and accuracy metrics. These components together support the construction and evaluation of the MIQA model (Section~\ref{sec:miqa_space}), forming the complete assessment pipeline. Section~\ref{sec:instantiations} details the task-specific implementations of these core components.

\subsection{Fundamental Components} \label{sec:paradigm}

\subsubsection{Data Space} \label{sec:input_space} 
Input original images $x \in \mathcal{X}$ undergo perturbations modeled via transformation operator $O: \mathcal{X} \times \mathcal{W} \times \Omega \to \mathcal{X}'$, where $\mathcal{W}$ denotes the perturbation parameter space and $\Omega$ represents the spatial weighting function that modulates perturbation intensity across the image domain. The perturbed image space $\mathcal{X}'$ comprises all images obtainable via these transformations.
  
\subsubsection{Vision Task and Model Space} \label{sec:model_task_space}
Let $\mathcal{Y}$ represent the task-specific response space associated with a machine vision task $\mathcal{T}$, which is executed by an ensemble of $n$ independent predictive models $\mathcal{M} = \{m_1, m_2, \ldots, m_n\}$, where each model $m_i$ is formalized as: 
%\begin{equation}
	$m_i: \mathcal{X} \to \mathcal{Y}, \quad i = 1, 2, \ldots, n.$
%\end{equation}

\subsubsection{Label Space}  \label{sec:label_space}
The label space $\mathcal{S}$ defines the quality of the perturbed image $x' \in \mathcal{X}'$ for task $\mathcal{T}$, quantifying the degree of degradation's impact on model inference performance. The quality score is derived by aggregating the degradation response of the task models $\mathcal{M}$, forming the mean machine opinion score (MMOS), a machine-centric counterpart to the traditional mean opinion score (MOS) used in human-centric IQA.

\noindent\textbf{Consistency Score}: Quantifies model's prediction stability under perturbations relative to original image outputs, reflecting invariance to distribution shifts regardless of original prediction correctness. For model $m_i$, consistency is defined as:
\begin{equation}
	C_{m_i}(x, x') = \mathbf{M}(m_i(x'), m_i(x)).
\end{equation}
The consistency score across all models is weighted by their contributions:
\begin{equation} \label{eq:cons}
	\mathbb{E}_C(x, x') = \sum_{i=1}^{n} \alpha_i C_{m_i}(x, x'),
\end{equation}
where $\mathbf{M}(\cdot)$ quantifies the agreement between $m_i(x')$ for the perturbed image and $m_i(x)$ for the original image, which depends on the task-specific evaluation protocol.

\noindent\textbf{Accuracy Score}: Further assesses whether the perturbed predictions remain aligned with the ground-truth $g^*(x)$, measuring the model's ability to preserve task-specific correctness under degradation.
\begin{equation}
	A_{m_i}(x, x') = \mathbf{M}(m_i(x'), g^*(x)).
\end{equation}
The accuracy across all models is weighted by their respective contributions:
\begin{equation} \label{eq:acc}
	\mathbb{E}_A(x, x') = \sum_{i=1}^{n} \alpha_i A_{m_i}(x, x').
\end{equation}  

\noindent\textbf{Composite Score}: The unified metric formulates a convex combination of consistency and accuracy to jointly reflect correctness and stability: 
\begin{equation}
	S(x', \mathcal{T}) = \lambda_1 \cdot \mathbb{E}_C(x, x') + \lambda_2 \cdot \mathbb{E}_A(x, x'),
\end{equation}
with weights $\lambda_1, \lambda_2 \in \left[0,1\right] $ satisfying $\lambda_1 + \lambda_2 = 1$ to balance their contributions. Special cases emerge when $\lambda_1 = 0$ (pure accuracy score) or $\lambda_2 = 0$ (pure consistency score).

\subsubsection{MIQA Objective and Evaluation} \label{sec:miqa_space}
\noindent\textbf{Objective}: MIQA aims to learn an optimal quality assessment function $Q^*: \mathcal{X'} \times \mathcal{T} \to \mathbb{R}$ that predicts task-specific quality scores for perturbed images. Given a function space $\mathcal{F} = \{Q_\theta: \mathcal{X'} \times \mathcal{T} \to \mathbb{R} \mid \theta \in \Theta\}$ parameterized by $\theta$, the optimization objective is:
%\begin{equation}
	$\theta^* = \arg\min_{\theta \in \Theta} \mathcal{L}(Q_\theta, \mathcal{D})$,
%\end{equation}
where $\mathcal{L}(\cdot, \cdot)$ measures the discrepancy between predicted and ground truth scores on dataset $\mathcal{D}$.

\noindent\textbf{Inference}: Given the optimal MIQA model $Q_{\theta^*}$, the quality prediction for perturbed image $x'$ under task $\mathcal{T}$ is: 
%\begin{equation}
	$\hat{S}(x', \mathcal{T}) = Q_{\theta^*}(x', \mathcal{T})$, 
%\end{equation}
where $\hat{S}(x', \mathcal{T})$ denotes the predicted quality score.

\noindent\textbf{Evaluation}: Given dataset $\mathcal{D}=\{(x'_i, S(x'_i,\mathcal{T}))\}_{i=1}^{N}$ with ground truth scores $S(x'_i,\mathcal{T})$, model performance is evaluated by:
%\begin{equation}
	$\mathbb{P} = \mathcal{E} \left(\{\hat{S}(x'_i, \mathcal{T})\}_{i=1}^{N}, \{S(x'_i, \mathcal{T})\}_{i=1}^{N} \right)$,
%\end{equation}
where $\mathcal{E}$ denotes evaluation metrics, such as Spearman's rank correlation (SRCC), Pearson's linear correlation (PLCC), and root mean squared error (RMSE).

\subsection{Task-Specific Instantiations}% and Implementation Details} 
\label{sec:instantiations} 
\subsubsection{Spatial Degradation Modes} \label{sec:spatial_modes}
The transformation operator generates a perturbed image $x' = O(x, w, \omega)$ by applying a transformation to the original image $x$ under parameter $w \in \mathcal{W}$ with spatial weighting $\omega \in \Omega$. To study the spatial distortion sensitivity of MVS, we define three canonical distortion modes (\eg, uniform distortion (UD), ROI-dominated distortion (ROI-DD), and background-dominated distortion (BG-DD)) by assigning spatially variant weights $\omega(p)$ to each pixel $p \in \mathcal{P}$:
\begin{equation}
	\omega(p) = 
	\begin{cases}
		\alpha, &p\in\mathcal{R} \\
		\beta, & p\notin\mathcal{R}
	\end{cases},
	\text{where} 
	\begin{cases}
		\alpha = \beta, & \text{for UD} \\
		\alpha > \beta, & \text{for ROI-DD} \\
		\alpha < \beta, & \text{for BG-DD}
	\end{cases}
\end{equation}
where $\mathcal{P}$ is the set of all pixel positions, $\mathcal{R}$ denotes ROI pixels, and $\alpha, \beta \in [0,1]$ determine the relative distortion strength applied to pixels inside and outside the ROI.

\subsubsection{Task-Specific Response Spaces and Model Weighting} \label{sec:task_response}
The structure of response space $\mathcal{Y}$ is task-dependent: For classification, $\mathcal{Y} = \{1, 2, \ldots, C\}$ for $C$ discrete classes. In object detection, $\mathcal{Y} = \{(b_1, c_1), \ldots, (b_k, c_k)\}$ where $b_i$ denotes bounding box coordinates and $c_i$ the corresponding class label. For segmentation, $\mathcal{Y} = \{1, 2, \ldots, C\}^{H \times W}$ where $H \times W$ denotes spatial dimensions with each pixel assigned a class label.  
Each model $m_i$ is associated with a weight $\alpha_i$ quantifying its contribution: $\sum_{i=1}^{n} \alpha_i = 1, \alpha_i \in [0, 1]$. The weights reflect each model's relative importance, derived from empirical performance or predefined criteria.
\begin{figure*}[t!]
	\centering  
	\includegraphics[width=0.99\textwidth]{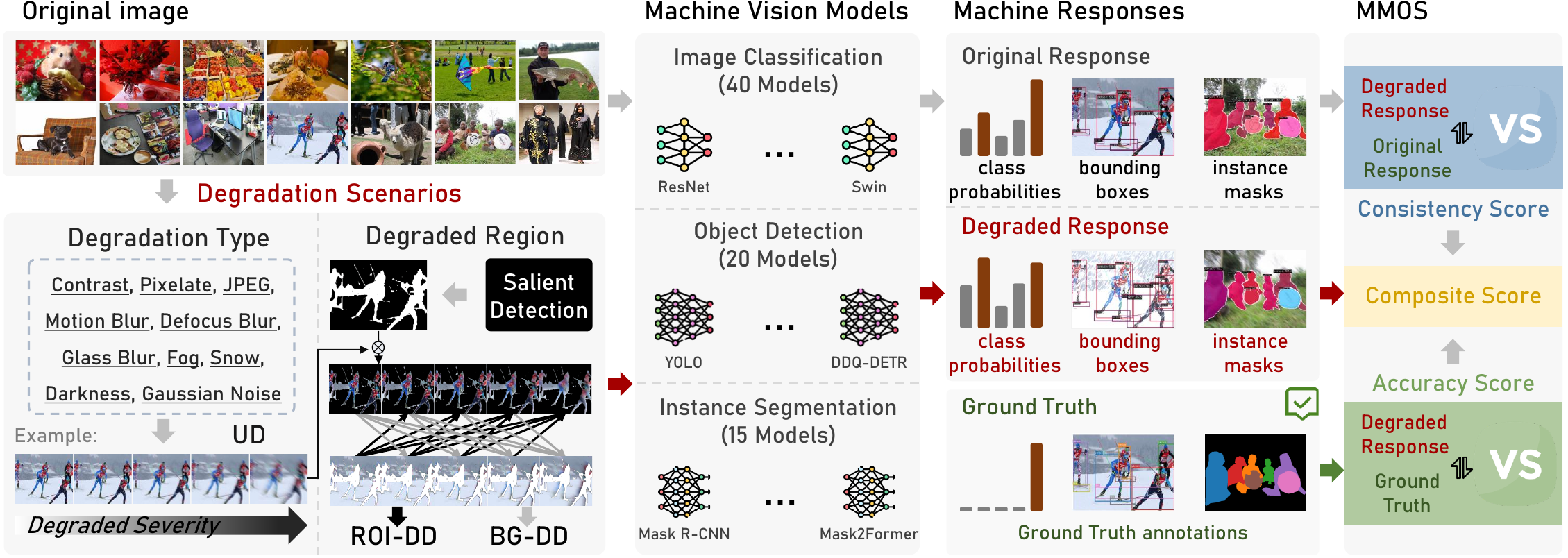}  \vspace{-.1em}
	\caption{Pipeline of the MIQD-2.5M construction.} \label{fig:pipeline} \vspace{-1.5em}
	
\end{figure*}
\subsubsection{Task-Specific Agreement Metrics} \label{sec:agreement_metrics}
\noindent\textbf{Image Classification}: The metric $\mathbf{M}$ is defined as:
\begin{equation}
	\mathbf{M}(y', y) = \mathbb{I}(y' = y), 
\end{equation}
where $y'$ is the prediction for perturbed image $x'$, and $y$ represents either ground truth $g^*(x)$ for accuracy metric or original prediction $m_i(x)$ for consistency metric. The indicator function $\mathbb{I}(\cdot)$ yields 1 if the condition holds true, 0 otherwise.

\noindent \textbf{Object Detection}: The metric $\mathbf{M}$ is instantiated through precision-recall analysis. Let $y' = \{(b_j', c_j')\}_{j=1}^{k'}$ denote the predicted bounding boxes and classes for perturbed input $x'$, and $y = \{(b_j, c_j)\}_{j=1}^k$ the reference annotations (ground truth or original predictions). Valid matches require an IoU above threshold $\tau$ and identical class labels. To ensure one-to-one correspondence, we employ greedy matching where each prediction and ground truth can participate in at most one matching pair. Under this constrained matching, precision and recall are formalized as: 
\begin{align} 
	&P_{m_i}(y', y) = \dfrac{\sum_{j'=1}^{k'} \mathbb{I}\left(\exists i \in \{1,\dots,k\}:M(b_{j'}', c_{j'}', b_i, c_i)\right)}{k'} , \notag\\
	&R_{m_i}(y', y)= \dfrac{\sum_{i=1}^{k} \mathbb{I}\left(\exists j' \in \{1,\dots,k'\}:M(b_{j'}', c_{j'}', b_i, c_i)\right)}{k}, \notag\\
	&M(b', c', b, c)= \left[\text{IoU}(b', b) \geq \tau \land c' = c\right].
\end{align}

Detection performance is summarized using mean Average Precision (mAP), averaged over all classes: $\dfrac{1}{C}\sum_{c=1}^{C} \int_0^1 P_c(r)dr$, where $P_c(r)$ represents precision at recall level $r$ for class $c$.

\noindent \textbf{Instance Segmentation}: Evaluation metrics retain the precision-recall framework of object detection but transition to mask-based criteria, both when evaluating accuracy against ground truth masks and when measuring consistency between predictions on original and perturbed images.
 
\section{The MIQD-2.5M Construction}\label{sec:database} 
This section details the construction pipeline~(as shown in Figure~\ref{fig:pipeline}), including modeling of uniform and region-specific degradations (Section~\ref{sec:data}); a collection of vision models spanning multiple tasks and architectures (Section~\ref{sec:model}); and in-depth analysis of model behaviors under varying degradation conditions (Section~\ref{sec:analysis}).
\begin{table*}[thb]\centering
	\caption{Summary of MIQD-2.5M characteristics for different machine tasks.} \vspace{-1em}
	\label{tab:dataset_stats}
	\resizebox{0.995\textwidth}{!}{
		%\large
		\begin{tabular}{c|c|c|c|c|c|c|c|c}
			\toprule 
			\textbf{Task}& \makecell{\textbf{Original} \\ \textbf{Image}} & \makecell{\textbf{Source of} \\ \textbf{Original image}} &\makecell{\textbf{Degradation} \\ \textbf{Type}} &\makecell{\textbf{Degradation} \\ \textbf{Severity}} & \makecell{\textbf{Degradation} \\ \textbf{Region type}} & \makecell{\textbf{Object} \\ \textbf{Category}} & \makecell{\textbf{Degraded} \\ \textbf{Image}}& \makecell{\textbf{Image} \\ \textbf{Resolution}} \\ 
			\midrule
			Image Classification & 5,000& ImageNet~\cite{imagenet} & 10 & 5 & 3 & 1,000 & 1,250,000&262$\times$415$\sim$4288$\times$2848  \\ \midrule
			\makecell{Object Detection \& \\ Instance Segmentation} & 5,000& MS COCO~\cite{mscoco}& 10 & 5 & 3 & 80 & 1,250,000 &200$\times$145$\sim$640$\times$640\\ \midrule 
			Total & 10,000& -&10 & 5 & 3 & $\geq$1,000 &2,500,000 &200$\times$145$\sim$4288$\times$2848 \\
			\bottomrule
		\end{tabular}
	} \vspace{-1em}
\end{table*}
\subsection{Data Acquisition and Degradation Modeling} \label{sec:data}
\noindent \textbf{Collection of Raw Images}:~To construct the MIQD-2.5M, we begin by sourcing high-quality raw images from well-established benchmark datasets. For image classification tasks, we utilize the ImageNet validation set~\cite{imagenet}, which comprises 50,000 images. These images are evaluated using two state-of-the-art IQA algorithms, HyperIQA~\cite{hyperiqa} and MANIQA~\cite{maniqa}. Each image is assigned a quality score by these IQA models, and the top 5 highest-rated images from each category are selected, ensuring a diverse and high-quality representation across various categories. For object detection and instance segmentation tasks, we select 5,000 images from the MS COCO validation set~\cite{mscoco}. Given the relatively limited size of the validation set, all validation images are subjected to degradation processing.
\begin{table}[t]\centering
	\caption{Summary of degradation regions, distortion intensity levels, and degradation categories. Each pair $(x,y)$ represents ROI distortion intensity $x$ and background distortion intensity $y$.} \vspace{-1em}
	\label{tab:distortions_stats}
	\resizebox{0.495\textwidth}{!}{
		\LARGE
		\begin{tabular}{cc|cc|cc|c|c}
			\toprule 
			\makecell{\textbf{Degraded} \\ \textbf{Region}} &  \makecell{\textbf{Degraded} \\ \textbf{Severity}}&	\makecell{\textbf{Degraded} \\ \textbf{Region}} &  \makecell{\textbf{Degraded} \\ \textbf{Severity}}&	\makecell{\textbf{Degraded} \\ \textbf{Region}} &  \makecell{\textbf{Degraded} \\ \textbf{Severity}} &\makecell{\textbf{Degraded} \\ \textbf{Category}}&\makecell{\textbf{Degraded} \\ \textbf{Name}}\\ 
			\midrule
			\multirow{10}{*}{\rotatebox{90}{\makecell{Uniform \\Distortion~(UD)}}}& \multirow{2}{*}{(1, 1)} &  	\multirow{10}{*}{\rotatebox{90}{\makecell{ROI-Dominated \\Distortion~(ROI-DD)}}}& (2, 1) &  \multirow{10}{*}{\rotatebox{90}{\makecell{Background-Dominated\\ Distortion~(BG-DD)}}}& (1, 2) &  \multirow{3}{*}{Digital}& Contrast \\ 
			& 	&  & (3, 1) &  & (1, 3) &  & Pixelate \\ 
			& \multirow{2}{*}{(2, 2)}  &  & (4, 1) &  & (1, 4) &  &  JPEG  \\ \cmidrule{7-8}
			& &	& (5, 1) & & (1, 5) &\multirow{3}{*}{Blur}& Motion blur\\ 
			& \multirow{2}{*}{(3, 3)} & 	&(3, 2) & &(2, 3) &  &  Defocus blur\\ 
			&  & & (4, 2) &  & (2, 4) &  &  Glass blur\\  
			\cmidrule{7-8} 
			&\multirow{2}{*}{(4, 4)} &	& (5, 2)  & & (2, 5)  &\multirow{3}{*}{\makecell{Environmental \\ Conditions}}& Fog \\ 
			& & &(4, 3)&  	&(3, 4)&  &  Snow \\  
			&  \multirow{2}{*}{(5, 5)}  &  &  (5, 3)  &&  (3, 5)  &  &  Darkness \\  \cmidrule{7-8} 
			&&&(5, 4)&&(4, 5)&\multirow{1}{*}{Noise}& Gaussian noise\\   
			\bottomrule
		\end{tabular}
	} \vspace{-2em}
\end{table}

\noindent \textbf{Degradation types and severity levels}:~To simulate diverse image degradation scenarios, each raw image undergoes systematic perturbations through ten distortion types organized into four categories:~\textbf{1)} Digital Distortions: Contrast variations, Pixelation, and JPEG compression artifacts each impact image quality by modifying intensity distributions, quantizing details into blocks, and introducing blocky artifacts, respectively.~\textbf{2)} Blur Distortions: Motion blur, defocus blur, and glass blur degrade sharpness by simulating camera movement, focus errors, or optical distortions.~\textbf{3)} Environmental conditions: Fog, snow, and darkness represent distinct environmental interference that degrades visual clarity, where fog causes light scattering, snow introduces particulate occlusion, and darkness reduces ambient illumination.~\textbf{4)} Noise Distortions: Gaussian noise characterized by random pixel variations, mimicking sensor noise or signal interference. Each distortion type is applied at five severity levels, ranging from subtle alterations to extreme impairments, ensuring a comprehensive spectrum of image quality variations under diverse conditions.
\begin{table}[t!]\centering
	\caption{Summary of the collected machine vision models across different architectures for image classification, object detection, and instance segmentation tasks.} \vspace{-1em}
	\label{tab:vision_models}
	\setcounter{rowno}{1}
	\resizebox{0.495\textwidth}{!}{
		\large
		\begin{tabular}{c|c|C|C}
			\toprule 
			\textbf{Task}&	\makecell{\textbf{Category} \\ \textbf{(Count)}} &\makecell{\textbf{Model} \\ \textbf{Name}} &\makecell{\textbf{Model} \\\textbf{Performance}}  \stepcounter{rowno}\\   
			\midrule % 
			\multirow{20}{*}{\rotatebox{90}{\makecell{Image Classification \\ (Top-1 Acc.~[\%] on ImageNet~\cite{imagenet})}}}&\multirow{7}{*}{\rotatebox{90}{\makecell{CNNs~\\(17)}}} &VGG19~\cite{vgg} &72.39 \stepcounter{rowno}\\  
			&&ResNet-34/50/10~\cite{resnet}1&76.42/80.39/82.32\stepcounter{rowno}\\ 
			&&Inception-v4~\cite{inceptionv4} &80.13 \stepcounter{rowno}\\   
			&&SE-ResNeXt-50/101~\cite{resnext} & 81.27/86.72 \stepcounter{rowno}\\ 
			&&SE-ResNet-50~\cite{squeeze}&81.10 \stepcounter{rowno}\\   
			&&EfficientNet-B2/B4/B6/B7~\cite{efficientnet}&80.31/83.26/84.79/86.85  \stepcounter{rowno}\\ 
			&&ConvNeXt-Tiny/Small/Base/Base$^\dagger$/Large~\cite{convnext}&82.05/83.14/83.85/85.81/87.46 \stepcounter{rowno}\\ 
			\cmidrule{2-4}
			%%%%%%%%%%%%%%%%%%%%%%%%%%%%%% 
			&\multirow{5}{*}{\rotatebox{90}{\makecell{ViTs\\(14)}}} & ViT-T-384/S-224/B-224/ViT-B-384/L-224~\cite{vit}&78.45/81.39/81.78/85.99/85.85\stepcounter{rowno}\\   
			&&Swin-T-224/S-224/B-384/L-224~\cite{swin}&81.39/83.23/84.48/86.32  \stepcounter{rowno}\\  
			&&DeiT-Tiny/Small/Base~\cite{deit}&72.16/79.85/81.98 \stepcounter{rowno}\\ 
			%&&DeiT-&\\ 
			&&CaiT-S-36-384~\cite{cait} &85.45\stepcounter{rowno}\\  
			&&MViTv2-Base~\cite{MViTv2} &84.44 \stepcounter{rowno}\\ 
			\cmidrule{2-4} 
			%%%%%%%%%%%%%%%%%%%%%%%%%%%%%%
			&\multirow{4}{*}{\rotatebox{90}{\makecell{Hybrids \\(5)}}}&CoaT-Small~\cite{coat}&82.36 \stepcounter{rowno}\\  
			&&CoAtNet$_1$~\cite{coatnet} &83.61 \stepcounter{rowno}\\ 
			&&LeViT-192/LeViT-384~\cite{LeViT}& 79.86/82.60  \stepcounter{rowno}\\ 
			&&Visformer-Small~\cite{Visformer}& 82.10 \stepcounter{rowno}\\ 
			\cmidrule{2-4}
			%%%%%%%%%%%%%%%%%%%%%%%%%%%%%%
			&\multirow{3}{*}{\rotatebox{90}{\makecell{Others\\ (4)}}}&
			MLP-Mixer-B-16-224~\cite{mixer}& 82.31 \stepcounter{rowno}\\  
			&&gMLP-S-16-224~\cite{gMLP}& 79.64 \stepcounter{rowno}\\ 
			&&MambaOut-Femto/MambaOut-Small~\cite{mambaout}& 79.85/84.50 \stepcounter{rowno}\\ 
			\midrule
			%%%%%%%%%%%%%%%%%%%%%%%%%%%%%%%%%%%%%%		
			\multirow{16}{*}{\rotatebox{90}{\makecell{Object Detection \\ (Box AP on MS COCO~\cite{mscoco})}}}&\multirow{6}{*}{\rotatebox{90}{\makecell{CNNs\\(10)}}} &Faster R-CNN (R-101-FPN)~\cite{faster_rcnn} &39.8\stepcounter{rowno}\\  
			&&Cascade R-CNN~(ResNet-101)~\cite{cascade} &42.0\stepcounter{rowno}\\ 
			&  &CenterNet~(ResNet-50)~\cite{centernet} &40.2 \stepcounter{rowno}\\
			&  &Sparse R-CNN~(ResNet-50)~\cite{sparse} &45.0\stepcounter{rowno}\\
			&  &YOLOX-L/YOLOX-X~\cite{yolox2021}&49.4/50.9 \stepcounter{rowno}\\
			&  &YOLO-V9-S/V10-M/V11-M/V11-L~\cite{Ultralytics_YOLO}&46.8/51.1/51.5/53.4 \stepcounter{rowno}\\ 	
			\cmidrule{2-4}
			&\multirow{5}{*}{\rotatebox{90}{\makecell{ViTs\\(5)}}} & PVT-Small~\cite{pvt} &40.4\stepcounter{rowno}\\ 
			&&Swin-Tiny~\cite{swin}& 46.0\stepcounter{rowno}\\
			&&ViTDet~(ViT-Base)~\cite{vitdet}&51.6\stepcounter{rowno}\\
			&&DINO~(Swin-L)~\cite{dino} &57.2\stepcounter{rowno}\\
			&&DDQ-DETR~(Swin-L)~\cite{ddq}&58.7\stepcounter{rowno}\\
			\cmidrule{2-4}
			&\multirow{5}{*}{\rotatebox{90}{\makecell{Hybrids\\(5)}}}& DETR~(ResNet-50)~\cite{detr}&39.9 \stepcounter{rowno}\\  
			&&Deformable DETR~(ResNet-50)~\cite{deformable_detr} &44.3 \stepcounter{rowno}\\ 
			&&DINO~(ResNet-50)~\cite{dino} &50.1 \stepcounter{rowno}\\
			&&DDQ-DETR~(ResNe-50)~\cite{ddq} &52.1 \stepcounter{rowno}\\
			&&CO-DETR~(ResNet-50) &54.8 \stepcounter{rowno}\\
			\midrule
			%%%%%%%%%%%%%%%%%%%%%%%%%%%%%%%%%%%%%%%
			\multirow{10}{*}{\rotatebox{90}{\makecell{Image Segmentation \\ (Mask AP on MS COCO~\cite{mscoco})}}}&\multirow{6}{*}{\rotatebox{90}{\makecell{CNNs\\(9)}}} &Mask R-CNN~(ResNet-50)~\cite{he2017mask} &37.1\stepcounter{rowno}\\  
			&&Cascade Mask R-CNN~(ResNet-101)~\cite{cascade_rcnn} &39.6\stepcounter{rowno}\\ 
			&  &PointRend~(ResNet-50)~\cite{pointrend} &38.0 \stepcounter{rowno}\\
			&&YOLO-V8-L/V9-E/V11-L/V11-X~\cite{Ultralytics_YOLO}& 42.6/44.3/42.9/43.8  \stepcounter{rowno}\\
			&& RTMDet-X~\cite{rtmdet}&44.6 \stepcounter{rowno}\\
			&& ConvNeXt-V2-Base~\cite{convnextv2}&46.4 \stepcounter{rowno}\\ 	\cmidrule{2-4}
			&\multirow{3}{*}{\rotatebox{90}{\makecell{ViTs\\(4)}}} &Swin-Tiny/Swin-Small~\cite{swin}& 41.7/43.2\stepcounter{rowno}\\  
			&& ViTDet~(ViT-Base)~\cite{vitdet}&45.7\stepcounter{rowno}\\
			&& Mask2Former (Swin-Small)~\cite{Mask2Former} &46.1\stepcounter{rowno}\\ 	\cmidrule{2-4} 
			&Hybrids~(2)& Mask2Former~(ResNet-50/ResNet-101)~\cite{Mask2Former}&42.9/44.0\stepcounter{rowno}\\  
			\bottomrule
		\end{tabular}
	} 
	\vspace{-1em}
\end{table}

\noindent \textbf{Degradation region and synthesis}:~Understanding distortion impacts across image regions is crucial as region relevance varies by task. We categorize distortions spatially: \textbf{1)} UD: Applied consistently across the entire image, providing a baseline for evaluating overall degradation effects. \textbf{2)} ROI-DD: Concentrates degradation on task-relevant regions, enabling analysis of how critical feature corruption affects performance. \textbf{3)} BG-DD: Applies stronger distortions to the background while preserving the ROI, simulating environmental variations that test model robustness to contextual changes. Our MIQD-2.5M generation process follows three steps: \textbf{a)} Apply uniform distortions as a baseline. \textbf{b)} Apply co-segmentation~\cite{coseg} to the original images to generate binary masks that encode the ROI, which are then used to separate the ROI from the background across all distortion severity levels. \textbf{c)} Recombine ROI and background regions with varying distortion intensities to create ROI-dominated and background-dominated degradations. For each distortion type, combining 5 severity levels for both ROI and background yields 10 unique combinations, plus the uniform distortions. This generates 250 degraded versions per raw image. The distortion characteristics of the datasets are summarized in Table~\ref{tab:distortions_stats}.

\noindent \textbf{Database statistics}:~Our database covers a wide range of distortion types, severity levels, and region-specific degradations. Table~\ref{tab:dataset_stats} summarizes key statistics, including the number of raw and distorted images, types of distortions, severity levels, degradation regions, and image resolutions. These datasets support the evaluation of diverse degradation scenarios and their impacts on machine vision, laying the foundation for understanding degradation mechanisms and developing effective MIQA models. 

\begin{figure*}[tb] \centering
	\includegraphics[width=0.325\textwidth]{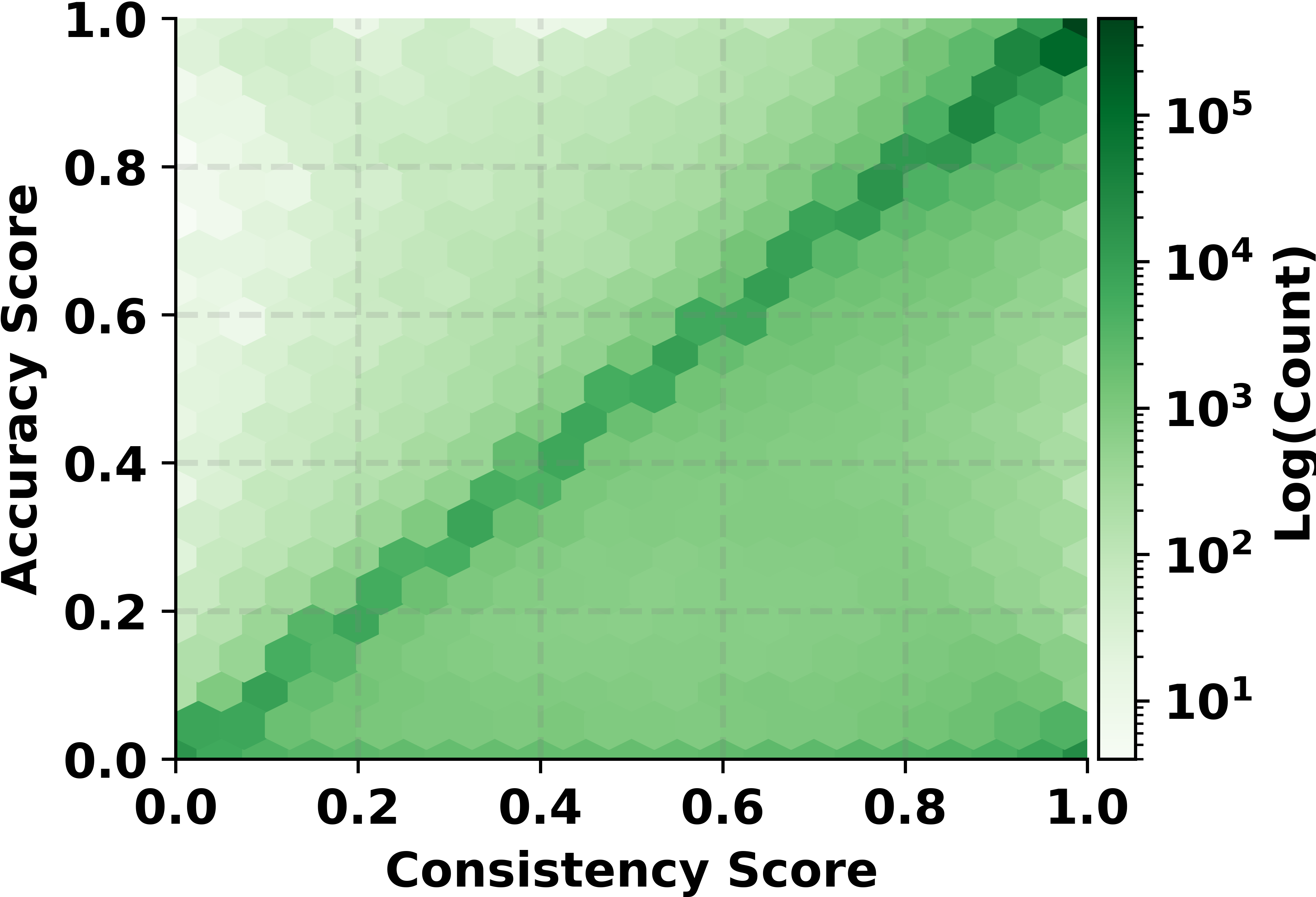}
	\includegraphics[width=0.325\textwidth]{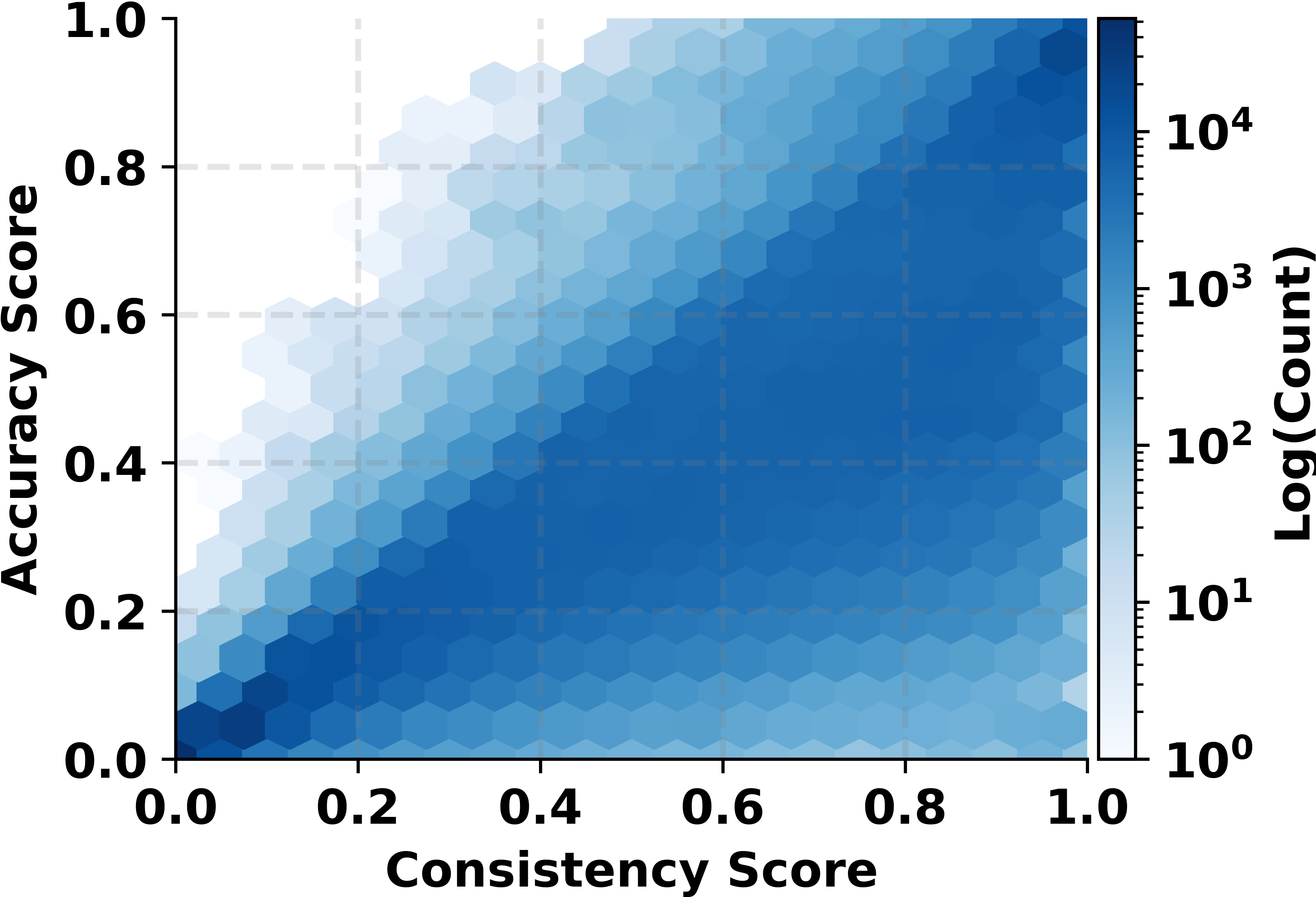}
	\includegraphics[width=0.325\textwidth]{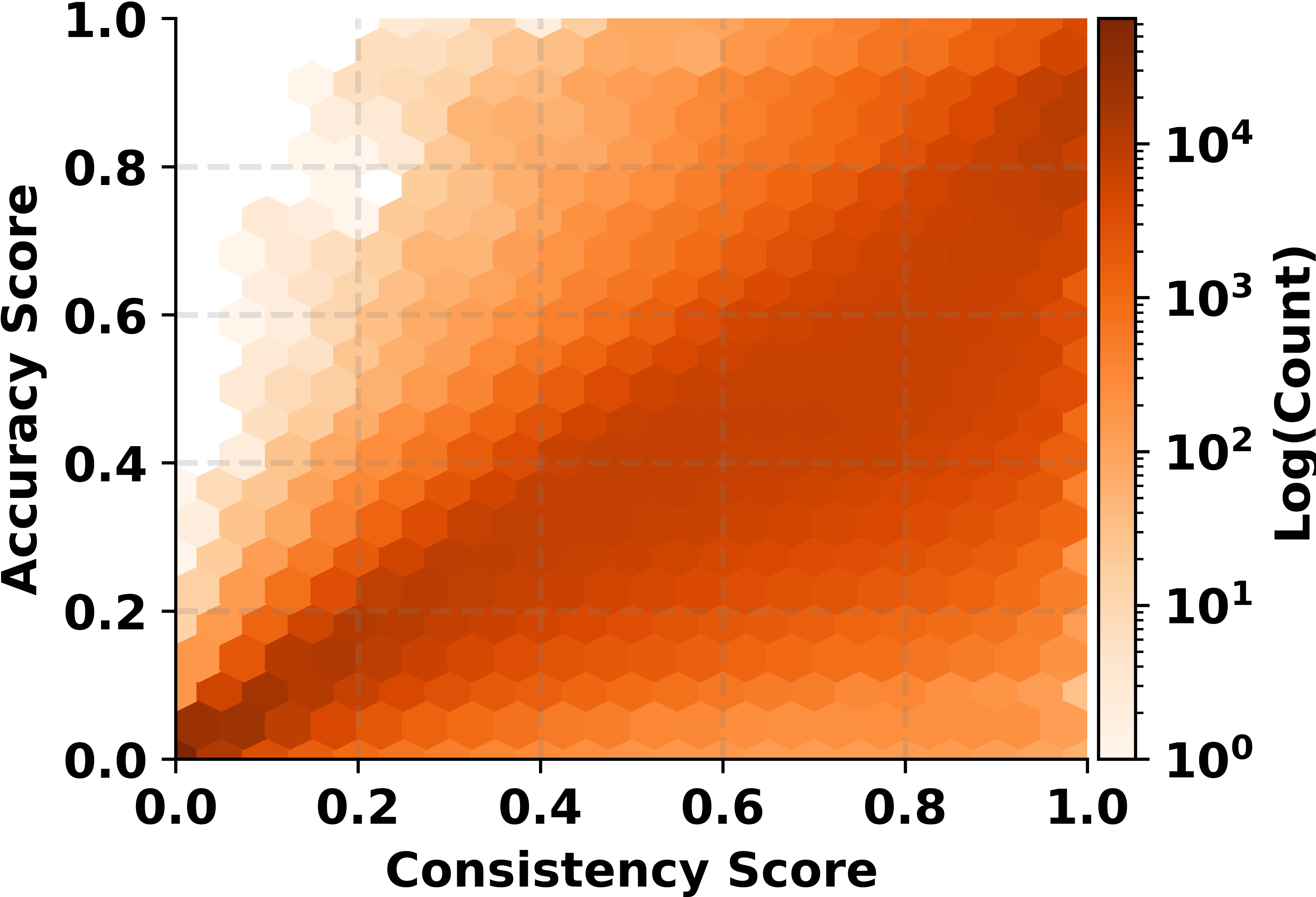}
	\caption{Density distribution of consistency and accuracy scores, where darker regions indicate higher sample concentration. From left to right: image classification, object detection, and instance segmentation tasks.} \label{fig:quality_distribution} \vspace{-1em}
\end{figure*}

\subsection{Collection of Machine Vision Models} \label{sec:model}
Table~\ref{tab:vision_models} presents a systematic collection of representative vision models across image classification, object detection, and instance segmentation tasks to analyze machine vision quality degradation. This diverse ensemble includes various architectural paradigms, such as CNNs, ViTs, CNN-ViT hybrids, and alternative models, with benchmark performance metrics serving as weighting coefficients for machine vision quality scores~(as outlined in Eqs.~\ref{eq:cons} and \ref{eq:acc}). The selection includes both canonical models (\eg, ResNet~\cite{resnet}, Faster R-CNN~\cite{faster_rcnn}, and Mask R-CNN~\cite{he2017mask}), which established foundational design principles, and advanced approaches (\eg, ConvNeXt~\cite{convnext}, Co-DETR~\cite{co-detr}, and MaskFormer~\cite{Mask2Former}). Additionally, architectures optimized for computational efficiency (\eg, EfficientNet~\cite{efficientnet} and YOLO~\cite{Ultralytics_YOLO} families) are incorporated. This model collection provides a solid empirical foundation for investigating quality degradation mechanisms across different architectural frameworks and visual tasks. We utilized these models, along with the degraded images, to generate consistency, accuracy, and composite scores according to the methodology outlined in Section~\ref{sec:formulations}. Then, we performed an analysis of the MIQD-2.5M.

\subsection{Analysis of MIQD-2.5M} \label{sec:analysis}
\begin{figure}[t!] \centering  
	\includegraphics[width=0.49\textwidth]{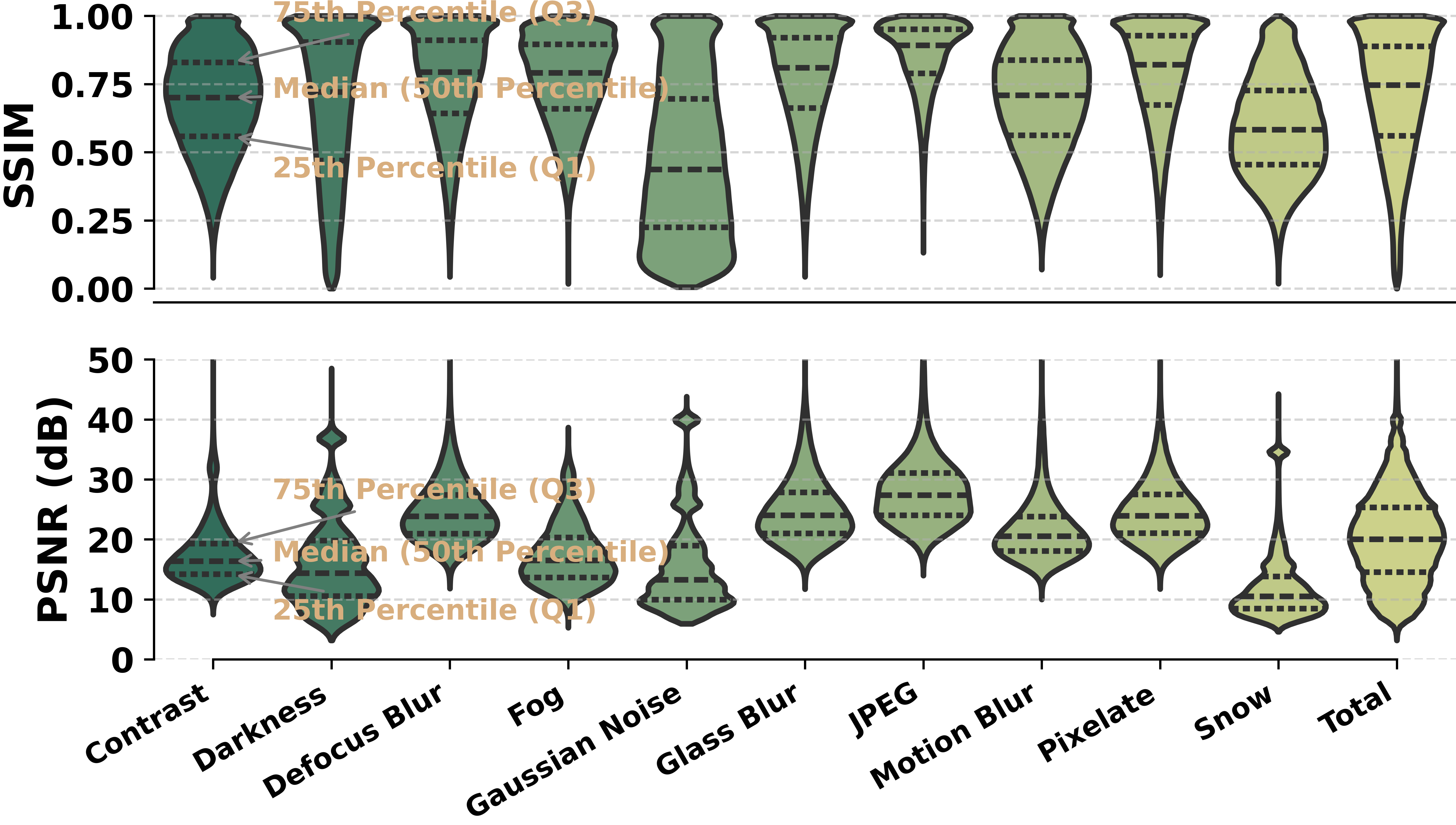}  \vspace{-1.em}
	\caption{Quality distribution of images under different distortion types, measured by SSIM and PSNR metrics.} \label{fig:dist_distribution} \vspace{-1.em}
\end{figure}

\noindent\textbf{Distribution of image distortion}:~The MIQA database presents a comprehensive range of distortions quantified by SSIM and PSNR metrics. As shown in Figure~\ref{fig:dist_distribution}, quartile analysis reveals distinctive degradation patterns across distortion types. Gaussian noise and snow artifacts exhibit lower median SSIM values with broader distributions, creating challenging test scenarios for model robustness. Conversely, JPEG compression and Pixelation show higher median quality metrics with narrower interquartile ranges, testing models' ability to detect subtle quality variations. The database spans from pristine to severely degraded samples, providing a comprehensive benchmarking platform for evaluating vision systems across diverse quality conditions.
\begin{figure*}[t] 
	\centering  
	\includegraphics[width=0.995\textwidth]{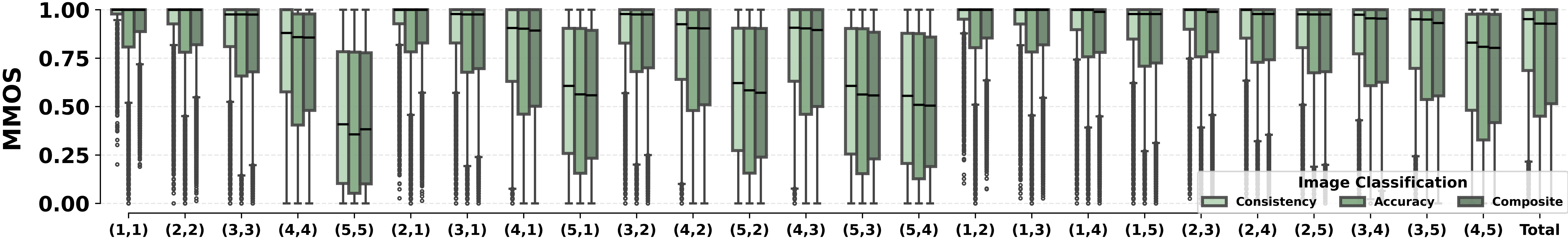}
	\includegraphics[width=0.995\textwidth]{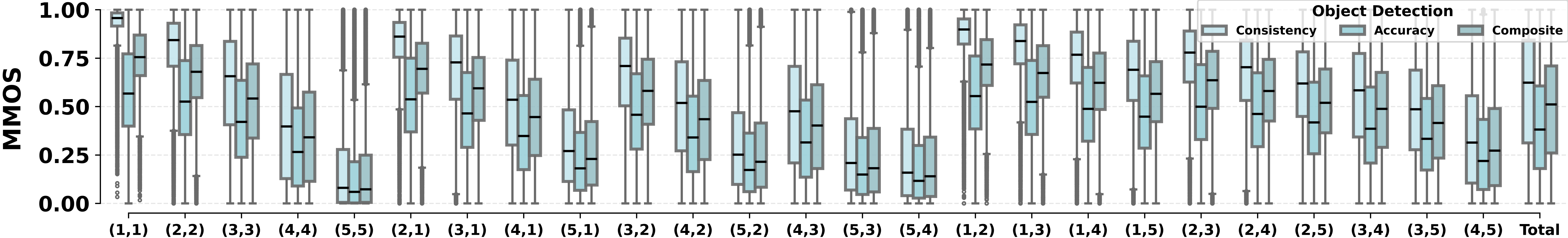} 
	\includegraphics[width=0.995\textwidth]{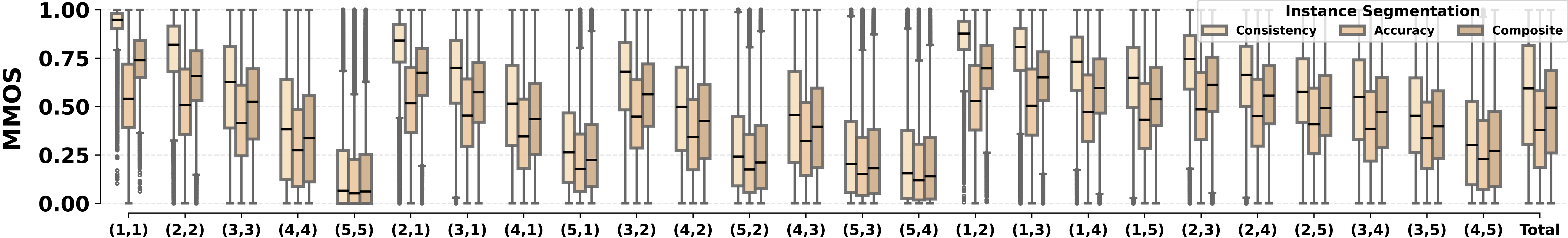}
	\caption{The distribution of MMOS values across different distortion regions and intensity levels in the three MIQA tasks.} \label{fig:metrics_distribution} \vspace{-1.em}
\end{figure*}

\noindent\textbf{Distribution of consistency and accuracy~scores}: Figure~\ref{fig:quality_distribution} displays density distributions across three vision tasks using consistency ($x$-axis) and accuracy ($y$-axis) scores. In classification, the distribution of consistency and accuracy scores predominantly follows a diagonal pattern, with additional samples dispersed throughout the entire sample space. For detection and segmentation tasks, a higher density is observed in the bottom-right quadrant, suggesting that consistency scores typically exceed accuracy scores. Specifically, samples with low consistency (\eg., $<$ 0.2) consistently demonstrate low accuracy. Across the majority of the samples, an increase in consistency score is accompanied by an improvement in accuracy, albeit with occasional exceptions where high consistency corresponds to poor accuracy. 

\noindent\textbf{Distribution of quality labels with distortion}:~Figure~\ref{fig:metrics_distribution} shows quality label distributions across visual tasks under varying distortion scenarios, using coordinates ($x,y$) to represent ROI and background distortion intensities. Analysis of UD ($x=y$), ROI-DD ($x>y$), and BG-DD ($x<y$) reveals distinct patterns in consistency, accuracy, and composite scores. Distortion effects follow a hierarchical pattern: increasing background distortion with constant foreground conditions causes a gradual performance decline, while increasing foreground distortion with constant background conditions leads to rapid deterioration. Simultaneous distortion in both regions compounds this degradation. This hierarchy confirms that critical region distortions primarily affect performance, while background perturbations mainly impair contextual reasoning. Under mild distortions, consistency scores exceed accuracy scores, demonstrating model resilience despite occasional low-quality predictions. As distortion intensifies, these metrics converge, indicating a fundamental compromise of system performance due to severe degradation of critical content.

\begin{table}[tb]\centering
	\caption{Statistical results from 100 cross-model label validation experiments. The best results are highlighted with colored backgrounds.} \vspace{-1.em}
	\label{tab:label_eval}
	\setcounter{rowno}{1}
	\resizebox{0.49\textwidth}{!}{
		\Large
		\begin{tabular}{c|C|CCC|CCC|CCC}
			\toprule 
			\multirow{2}{*}{\textbf{Metric}}&\multirow{2}{*}{\textbf{Statistic}}& \multicolumn{3}{c|}{\textbf{Image Classification}}& \multicolumn{3}{c|}{\textbf{Object Detection}}& \multicolumn{3}{c}{\textbf{Instance Segmentation}} \\   
			\cmidrule{3-11}
			& & \textbf{SRCC}$\uparrow$ & \textbf{PLCC}$\uparrow$ &\textbf{RMSE}$\downarrow$ &\textbf{SRCC}$\uparrow$ & \textbf{PLCC}$\uparrow$  &\textbf{RMSE}$\downarrow$  &\textbf{SRCC}$\uparrow$ & \textbf{PLCC}$\uparrow$  &\textbf{RMSE}$\downarrow$  \stepcounter{rowno} \\   \midrule 
			\textbf{Consistency}&Mean&0.867&0.926&0.110&0.942&0.948&0.103&0.941&0.947&0.101 \stepcounter{rowno} \\
			\textbf{Score}	&Std.&0.010&0.006&0.006&0.010&0.011&0.010&0.009&0.008&0.008 \stepcounter{rowno} \\ 	\cmidrule{1-11}
			\textbf{Accuracy}&Mean&0.913&\cellcolor{OliveGreen!20}0.961&0.103&0.958&\cellcolor{blue!20}0.961&\cellcolor{blue!20}0.078&\cellcolor{orange!20}0.959&\cellcolor{orange!20}0.961&\cellcolor{orange!20}0.074  \stepcounter{rowno}\\
			\textbf{Score}	&Std.&0.008&0.004&0.006&0.012&0.011&0.009&0.008&0.008&0.007  \stepcounter{rowno}\\ 	\cmidrule{1-11}
			\textbf{Composite}&Mean&\cellcolor{OliveGreen!20}0.914&0.952&\cellcolor{OliveGreen!20}0.095&\cellcolor{blue!20}0.959&0.960&0.081&\cellcolor{orange!20}0.959&0.960&0.077  \stepcounter{rowno}\\
			\textbf{Score}	&Std.&0.008&0.005&0.006&0.011&0.011&0.009&0.008&0.007&0.007  \stepcounter{rowno}\\
			\bottomrule
		\end{tabular}
	}  \vspace{-1.em}
\end{table}  
\noindent\textbf{Model bias and label validation}: 
%To assess the stability of the quality labels, we randomly partition the vision models into 80\% and 20\% subsets, both used to independently generate labels with the labeling method. This process is repeated over 100 random trials. 
To assess label stability, we randomly partition vision models into 80\% and 20\% subsets to independently generate labels, repeated over 100 trials. We then compute the correlation (SRCC, PLCC) and error (RMSE) between the two sets of labels for each task, as shown in Table~\ref{tab:label_eval}. Classification models demonstrate substantial inter-model agreement with SRCC values exceeding 0.86, while detection and segmentation tasks achieve even stronger consensus with correlations above 0.94. The distinct patterns between SRCC and PLCC values in classification reveal sophisticated non-linear quality relationships that capture nuanced degradation effects beyond simple monotonic trends. Across all tasks, accuracy scores consistently achieve superior performance with higher correlation values, demonstrating that models exhibit coordinated and predictable responses when evaluated against ground truth standards. This coordinated behavior indicates that as image quality deteriorates, models maintain systematic patterns in their adherence to or deviation from ground truth, showcasing the fundamental robustness principles underlying modern vision architectures. The composite score that integrates both consistency and accuracy dimensions consistently delivers higher correlations and lower RMSE across all tasks, establishing it as the optimal metric for comprehensive quality assessment. These results demonstrate effective labeling methodology through the relatively strong cross-model agreement, validating the generalizability of the MMOS labels across vision models within each task domain.
 
\section{The Proposed MIQA Method} \label{sec:method}
\begin{figure}[t!] \centering
	\includegraphics[width=0.485\textwidth]{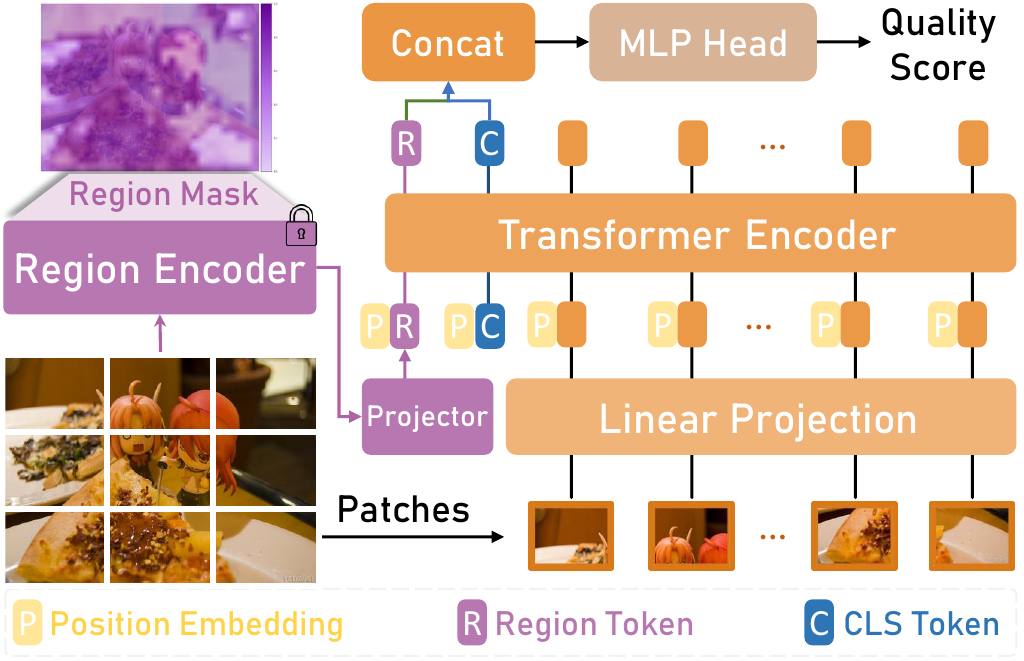} \vspace{-1.em}
	\caption{Structure of the proposed RA-MIQA model.} \vspace{-1.em}
	\label{fig:model}
\end{figure}

\noindent\textbf{Model overview}: We propose RA-MIQA, a model that implements a mapping $\mathcal{Q}: \mathbb{R}^{3 \times H \times W} \rightarrow \mathbb{R}$ from an input image to its quality score. As shown in Figure~\ref{fig:model}, RA-MIQA comprises three key components: a frozen region encoder that extracts a distortion region-aware token, a Transformer encoder that models interactions among patch, class, and region tokens, and a regression head that integrates region and class representations for final prediction.%We propose RA-MIQA that realizes a mapping $\mathcal{Q}: \mathbb{R}^{3 \times H \times W} \rightarrow \mathbb{R}$ from an input image $\mathbf{x}$ to its quality score. As shown in Figure~\ref{fig:model}, RA-MIQA consists of three components: a frozen pretrained region encoder that generates a region token representing distortion-sensitive areas, a Transformer encoder that facilitates interaction between patch embeddings, class, and region tokens, and a regression head that fuses region and class tokens for quality prediction.
%As shown in Figure~\ref{fig:model}, RA-MIQA integrates three principal components: a frozen pretrained region encoder generating distortion region-sensitive representation, a vision Transformer facilitating token interactions, and a regression head for quality prediction.

%\subsection{Feature Extraction}

\noindent\textbf{Patch and region embedding}: The input image ${x}$ is processed via two parallel branches. For the patch embedding, ${x}$ is divided into $N_p = \frac{H \times W}{P^2}$ non-overlapping patches, each linearly projected into a $D$-dimensional embedding space:
%\begin{equation}
	$\mathbf{Z}_p = \mathcal{P}({x}) \in \mathbb{R}^{N_p \times D}.$
%\end{equation}
Concurrently, a frozen pretrained region encoder $\mathcal{R}$~\cite{gao2022luss} extracts spatial feature maps:
\begin{equation}
	\mathbf{F}_r = \mathcal{R}({x}) \in \mathbb{R}^{H_r \times W_r \times D_r},
\end{equation}
where $H_r$,~$W_r$, and~$D_r$ denote the feature height, width, and dimension, respectively. This encoder remains frozen during training, ensuring performance gains stem purely from its information content rather than additional parameter optimization.  Average pooling then yields a compact representation~$\bar{\mathbf{F}}_r \in \mathbb{R}^{D_r}$, projected into the embedding space:
\begin{equation}
	\mathbf{Z}_r = \mathcal{G}(\bar{\mathbf{F}}_r) = \mathrm{LN}(\phi(\mathbf{W}_r \bar{\mathbf{F}}_r + \mathbf{b}_r)) \in \mathbb{R}^D,
\end{equation}
where $\mathbf{W}_r \in \mathbb{R}^{D \times D_r}$, $\mathbf{b}_r \in \mathbb{R}^D$, $\mathrm{LN}$ denotes layer normalization, and $\phi(\cdot)$ represents the GELU activation.

\noindent\textbf{Sequence construction}: We construct a sequence by combining the region token $\mathbf{Z}_r \in \mathbb{R}^D$, a learnable class token $\mathbf{Z}_c \in \mathbb{R}^D$, and patch embeddings:
%\begin{equation}
$	\mathbf{Z}_0 = [\mathbf{Z}_r; \mathbf{Z}_c; \mathbf{Z}_p] \in \mathbb{R}^{(N_p+2) \times D}.$
%\end{equation}
Position embeddings $\mathbf{E}_{\text{pos}} \in \mathbb{R}^{(N_p+2) \times D}$ are added to incorporate spatial information:
%\begin{equation}
$	\mathbf{Z}_0' = \mathbf{Z}_0 + \mathbf{E}_{\text{pos}}.$
%\end{equation}

\noindent\textbf{Transformer processing}: This sequence $\mathbf{Z}_0'$ passes through $L$ transformer blocks, each comprising multi-head self-attention (MSA) and feed-forward networks (FFN):
\begin{equation}
\begin{aligned}
	\mathbf{Z}_{\ell}' &= \text{MSA}(\text{LN}(\mathbf{Z}_{\ell-1})) + \mathbf{Z}_{\ell-1}, \\
	\mathbf{Z}_{\ell} &= \text{FFN}(\text{LN}(\mathbf{Z}_{\ell}')) + \mathbf{Z}_{\ell}',
\end{aligned}
\end{equation}
for $\ell=1,\ldots,L$. This architecture enables rich interactions between class token, region token, and patch embeddings, capturing relationships between global image characteristics and localized distortions. After processing, we extract the final representations:
\begin{equation}
	\mathbf{h}_r = \mathbf{Z}_{L}[0,:] \in \mathbb{R}^{D},~
	\mathbf{h}_c = \mathbf{Z}_{L}[1,:] \in \mathbb{R}^{D}.
\end{equation}

%\subsection{Quality Regression}
\noindent\textbf{Quality regression}:
The quality score $\hat{s}$ is predicted by concatenating region and class token representations and processing through a regression head:
\begin{equation}
	%q~=~\mathcal{H}([\mathbf{h}_r; \mathbf{h}_c])~=~\mathbf{W}_2(\phi(\mathbf{W}_1[\mathbf{h}_r; \mathbf{h}_c] + \mathbf{b}_1)) + \mathbf{b}_2~\in~\mathbb{R},
	\hat{s}~=~\mathbf{W}_2(\phi(\mathbf{W}_1[\mathbf{h}_r; \mathbf{h}_c] + \mathbf{b}_1)) + \mathbf{b}_2~\in~\mathbb{R},
\end{equation}
where $\mathbf{W}_1 \in \mathbb{R}^{D \times 2D}$, $\mathbf{b}_1 \in \mathbb{R}^{D}$, $\mathbf{W}_2 \in \mathbb{R}^{1 \times D}$, and $\mathbf{b}_2 \in \mathbb{R}$ are learnable parameters. This fusion combines global image context (via class token) with distortion-specific information (via region token) for accurate quality assessment.
\section{Experiment}
\label{sec:benchmark}
\subsection{Experimental Setup}  
\noindent\textbf{Dataset partitioning}: The dataset (summarized in Table~\ref{tab:dataset_stats}) employs a 4:1 ratio to divide original images, with corresponding degraded versions allocated to training and validation sets respectively, preventing content leakage between partitions. To address class imbalances, we implement task-specific stratified sampling. For image classification, stratification preserves the 4:1 ratio within each class category. For object detection and instance segmentation, we first apply class-based stratified random sampling to single-class images, then dynamically allocate multi-class images based on real-time class proportions in the training set, prioritizing underrepresented categories. Final adjustments ensure balanced class distributions across both partitions.

\noindent\textbf{Baselines and evaluation criteria}: 
We first evaluated the applicability of both HVS-based FR-IQA metrics (\ie, PSNR, SSIM~\cite{ssim}, VSI~\cite{vsi}, LPIPS~\cite{lpips}, and DISTS~\cite{dists}) and NR-IQA methods (\ie, HyperIQA~\cite{hyperiqa} and MANIQA~\cite{maniqa} pre-trained on KonIQ-10k~\cite{koniq10k}) for machine-based IQA tasks. Then, we benchmarked representative neural architectures spanning CNNs and ViTs, including both standard and lightweight variants: ResNet-18~\cite{resnet}, ResNet-50~\cite{resnet}, EfficientNet-b1~\cite{efficientnet}, EfficientNet-b5~\cite{efficientnet}, and ViT-small~\cite{vit}, for training and validation on MIQA tasks. The proposed RA-MIQA is built upon the ViT-small model with further improvements. To ensure a fair comparison, the predicted scores were further nonlinearly regressed to align with the ground-truth using a five-parameter logistic function of the form:
\begin{equation}
	f({q}) = \alpha_1 \left( \frac{1}{2} - \frac{1}{1 + e^{\alpha_2({q} - \alpha_3)}} \right) + \alpha_4 {q} + \alpha_5,
\end{equation}
where $\alpha_1$ to $\alpha_5$ are parameters fitted on the validation set. The performance was evaluated using multiple correlation metrics: SRCC, PLCC, and Kendall's Rank Correlation Coefficient (KRCC), all ranging from 0 to 1 with higher values indicating better performance, alongside RMSE to quantify prediction accuracy across diverse measurement methods. 
\begin{table*}[thb]\centering
	\caption{Comparison of HVS-based and machine-based IQA methods on the composite score. The best results within each category are highlighted with a colored background. The last row reports the gains relative to ViT-small.} \vspace{-1.em}
	\label{tab:resutls} 
	\setcounter{rowno}{1}
	\resizebox{0.99\textwidth}{!}{
		%\large   \rowcolor{white}
		\begin{tabular}{c|C|CCCC|CCCC|CCCC}
			\toprule 
			\multirow{2}{*}{\textbf{Category}}&	\multirow{2}{*}{\textbf{Method}}& \multicolumn{4}{c|}{\textbf{Image Classification}}& \multicolumn{4}{c|}{\textbf{Object Detection}}& \multicolumn{4}{c}{\textbf{Instance Segmentation}}  \\   
			\cmidrule{3-14}
			& &\textbf{SRCC}$\uparrow$ & \textbf{PLCC}$\uparrow$  & \textbf{KRCC}$\uparrow$  &\textbf{RMSE}$\downarrow$ &\textbf{SRCC}$\uparrow$ & \textbf{PLCC}$\uparrow$  & \textbf{KRCC}$\uparrow$  &\textbf{RMSE}$\downarrow$  &\textbf{SRCC}$\uparrow$ & \textbf{PLCC}$\uparrow$  & \textbf{KRCC}$\uparrow$  &\textbf{RMSE}$\downarrow$ \stepcounter{rowno}\\   \midrule 
			
			\multirow{8}{*}{\rotatebox{90}{HVS-based}}&PSNR&0.2388 &0.2292&0.1661&0.2928&0.3176&0.3456 & 0.2148 	& 0.2660 &0.3242 & 0.3530  & 0.2196 &0.2553 \stepcounter{rowno}\\
			&SSIM~\cite{ssim}&0.3027&0.2956& 0.2119 &0.2874	& 0.4390	& 0.4505&0.3011	& 0.2531 &0.4391&0.4512&0.3011 & 0.2435    \stepcounter{rowno}\\
			&VSI~\cite{vsi}&0.3592&0.3520&0.2520 &0.2816&0.4874 	&0.4940&0.3355	&0.2465&0.4919 &  0.4985&0.3392	&0.2365 \stepcounter{rowno}\\
			&LPIPS~\cite{lpips}	&0.3214 &0.3280&0.2258&0.2842&0.5264 &\cellcolor{blue!20}0.5376 &\cellcolor{blue!20}0.3697&\cellcolor{blue!20}0.2390&0.5342&\cellcolor{orange!20}0.5453&\cellcolor{orange!20}0.3754&\cellcolor{orange!20}0.2287 \stepcounter{rowno}\\  
			&DISTS~\cite{dists}	&\cellcolor{OliveGreen!20}0.3878&\cellcolor{OliveGreen!20}0.3804&\cellcolor{OliveGreen!20}0.2724 &\cellcolor{OliveGreen!20}0.2782&\cellcolor{blue!20}0.5266 &0.5352&0.3659&0.2395&\cellcolor{orange!20}0.5363&0.5450 &0.3738 &0.2288 \stepcounter{rowno}\\ \cdashline{2-14} \addlinespace[3pt] 
			&HyperIQA~\cite{hyperiqa}&0.2496& 0.2279& 0.1741 & 0.2929&0.4462&0.4463& 0.3031&0.2537&0.4456& 0.4518 &0.3031& 0.2434 \stepcounter{rowno}\\ 
			&MANIQA~\cite{maniqa}&0.3403& 0.3255& 0.2387&0.2844& 0.4574& 0.4617&0.3124& 0.2515 &0.4636&0.4680&0.3176&0.2411 \stepcounter{rowno}\\  \midrule
			\multirow{7}{*}{\rotatebox{90}{Machine-based}}%&HyperIQA$^*$&-&-	&-	& - &-	& -&-	& -&-	& -&-	& -&-	& -\\ 
			&ResNet-18~\cite{resnet} &0.5131	&0.5427& 0.3715 &0.2527	&0.7541&0.7734& 0.5625&0.1797& 0.7582&0.7790&0.5674&0.1711 \stepcounter{rowno}\\
			&ResNet-50~\cite{resnet} &0.5581	& 0.5797&0.4062& 0.2451&0.7743	&0.7925&0.5824&0.1729&0.7729&0.7933&0.5826&0.1661 \stepcounter{rowno}\\
			&EfficientNet-b1~\cite{efficientnet}&0.5901&0.6130&0.4320 &0.2377&0.7766&0.7950	&0.5859&0.1720	&0.7808&0.7999	&0.5918&0.1637	\stepcounter{rowno} \\% \cmidrule{2-14}
			&EfficientNet-b5~\cite{efficientnet}&0.6330&0.6440&0.4680&0.2301&0.7866&0.8041&0.5971& 0.1685& 0.7899&0.8074&0.6013&0.1610 \stepcounter{rowno} \\ 
			&ViT-small~\cite{vit}&0.5998&0.6161&0.4407&0.2370	& 0.7992&0.8142	& 0.6099&0.1646	&0.7968&0.8139	& 0.6083&0.1585 \stepcounter{rowno}\\
			\cmidrule{2-14} 
			
			&{RA-MIQA}&\cellcolor{OliveGreen!20}\makecell{0.7003 \\ \textcolor{red!100}{\textbf{+16.76\%}}}&\cellcolor{OliveGreen!20}\makecell{0.6989 \\ \textcolor{red!100}{\textbf{+13.44\%}}}&\cellcolor{OliveGreen!20}\makecell{0.5255 \\ \textcolor{red!100}{\textbf{+19.24\%}}}&\cellcolor{OliveGreen!20}\makecell{0.2152 \\ \textcolor{red!100}{\textbf{+9.20\%}}}&\cellcolor{blue!20}\makecell{0.8125 \\ \textcolor{red!100}{\textbf{+1.66\%}}}&\cellcolor{blue!20}\makecell{0.8264 \\ \textcolor{red!100}{\textbf{+1.50\%}}}&\cellcolor{blue!20}\makecell{0.6263 \\ \textcolor{red!100}{\textbf{+2.69\%}}}&\cellcolor{blue!20}\makecell{0.1596 \\ \textcolor{red!100}{\textbf{+3.04\%}}}&\cellcolor{orange!20}\makecell{0.8188 \\ \textcolor{red!100}{\textbf{+2.76\%}}}&\cellcolor{orange!20}\makecell{0.8340 \\ \textcolor{red!100}{\textbf{+2.47\%}}}&\cellcolor{orange!20}\makecell{0.6333 \\ \textcolor{red!100}{\textbf{+4.11\%}}}&\cellcolor{orange!20}\makecell{0.1505 \\ \textcolor{red!100}{\textbf{+5.05\%}}} \stepcounter{rowno}\\
			\bottomrule  
		\end{tabular} 
	}   \vspace{-.7em}
\end{table*}  
\begin{table*}[t]
	\centering
	\caption{Comparison between HVS-based and machine-based IQA methods on consistency and accuracy scores. The best results within each category are highlighted with a colored background. The last row reports the gains relative to ViT-small.} \vspace{-1.em}
	\label{tab:cons_acc}
	\setcounter{rowno}{1}
	\resizebox{0.99\textwidth}{!}{
		%\large 
		\begin{tabular}{c|CCC|CCC|CCC|CCC|CCC|CCC}
			\toprule 
			\multirow{3}{*}{\textbf{Method}}& \multicolumn{6}{c|}{\textbf{Image Classification}}&\multicolumn{6}{c|}{\textbf{Object Detection}} &\multicolumn{6}{c}{\textbf{Instance Segmentation}}  \\   \cmidrule{2-19} 
			\tiny &\multicolumn{3}{c|}{\textbf{Accuracy Score}}  &\multicolumn{3}{c|}{\textbf{Consistency Score}} &\multicolumn{3}{c|}{\textbf{Accuracy Score}}  &\multicolumn{3}{c|}{\textbf{Consistency Score}} &\multicolumn{3}{c|}{\textbf{Accuracy Score}}  &\multicolumn{3}{c}{\textbf{Consistency Score}} \\  
			\tiny &\textbf{SRCC}$\uparrow$ & \textbf{PLCC}$\uparrow$&\textbf{RMSE}$\downarrow$ &\textbf{SRCC}$\uparrow$ & \textbf{PLCC}$\uparrow$&\textbf{RMSE}$\downarrow$ &\textbf{SRCC}$\uparrow$ & \textbf{PLCC}$\uparrow$&\textbf{RMSE}$\downarrow$ &\textbf{SRCC}$\uparrow$ & \textbf{PLCC}$\uparrow$&\textbf{RMSE}$\downarrow$&\textbf{SRCC}$\uparrow$ & \textbf{PLCC}$\uparrow$&\textbf{RMSE}$\downarrow$ &\textbf{SRCC}$\uparrow$ & \textbf{PLCC}$\uparrow$&\textbf{RMSE}$\downarrow$\stepcounter{rowno}\\   \midrule 
			PSNR&0.2034 &  0.1620&0.3541&0.2927 &0.2812&0.2692&0.2234 & 0.2449 &0.2747& 0.3712  &0.3933&0.2839& 0.2182&0.2398&0.2616&0.3796&0.4061& 0.2770  \stepcounter{rowno}\\
			SSIM~\cite{ssim}&0.2529&0.2101& 0.3509&  0.3740 &0.3663 & 0.2610&0.3434 &0.3419& 0.2662& 0.5128 &0.5130 &0.2651& 0.3271 &0.3284 &0.2545&0.5174 &0.5204 &0.2589  \stepcounter{rowno}\\
			VSI~\cite{vsi}&0.3020&0.2515&0.3473& 0.4392 &0.4336& 0.2528&0.3799&0.3685& 0.2634& 0.5700 &0.5571&0.2565&0.3703&0.3645&0.2509&0.5757&0.5749&0.2481 \stepcounter{rowno} \\ 
			LPIPS~\cite{lpips}&0.2680 &	0.2355 &0.3488&0.3927 &0.4032  & 0.2567& 0.4064&0.3987& 0.2598 &\cellcolor{blue!20}0.6196 &\cellcolor{blue!20}0.6232&\cellcolor{blue!20} 0.2415&0.3972&0.3941&0.2476&\cellcolor{orange!20}0.6300 &\cellcolor{orange!20}0.6344&\cellcolor{orange!20} 0.2344 \stepcounter{rowno}\\
			DISTS~\cite{dists}&\cellcolor{OliveGreen!20}0.3291&\cellcolor{OliveGreen!20} 0.2768 &\cellcolor{OliveGreen!20}0.3448&\cellcolor{OliveGreen!20}0.4683&\cellcolor{OliveGreen!20} 0.4628& \cellcolor{OliveGreen!20}0.2487&\cellcolor{blue!20}0.4089&\cellcolor{blue!20}0.3999& \cellcolor{blue!20}0.2597&0.6174&0.6178&0.2429&\cellcolor{orange!20}0.4069&\cellcolor{orange!20}0.4012&\cellcolor{orange!20}0.2468 &0.6255 &0.6270 & 0.2362  \stepcounter{rowno} \\ \cdashline{1-19} \addlinespace[3pt]
			HyperIQA~\cite{hyperiqa}&0.2100 &0.1649 & 0.3540& 0.2966& 0.2777 & 0.2695&0.3646& 0.3545& 0.2649 & 0.5009& 0.4943& 0.2684&0.3486&0.3442&0.2530&	0.5056&0.4995& 0.2626 \stepcounter{rowno}  \\
			MANIQA~\cite{maniqa}&0.2924&0.2435& 0.3481&0.3963&0.3870 & 0.2587& 0.3839& 0.3823&0.2618&0.4991&0.4975 & 0.2679 &0.3755&0.3749&0.2498&0.5096& 0.5098&  0.2608  \stepcounter{rowno} \\  	\midrule 
			ResNet-50~\cite{resnet}&0.4734&0.4411 &0.3221& 0.5989& 0.6551& 0.2119&0.6955& 0.6898& 0.2051& 0.8252& 0.8457& 0.1648 &0.6863&0.6847 &0.1964&0.8320& 0.8480 &0.1607 \stepcounter{rowno}\\
			EfficientNet-b5~\cite{efficientnet}&0.5586& 0.5149& 0.3076&0.6774&0.7168& 0.1956&0.7042& 0.6991 &0.2026&0.8353& 0.8530& 0.1612 &0.6933&0.6949&0.1938&0.8419&0.8564& 0.1565 \stepcounter{rowno}\\
			ViT-small~\cite{vit}& 0.5788& 0.5197 & 0.3066&0.6798& 0.7189 &0.1950 &0.7121&0.7052&  0.2008 & 0.8459& 0.8620 &0.1566&0.7168&0.7146& 0.1885& 0.8487& 0.8616&  0.1539 \stepcounter{rowno}\\   
			\cmidrule{1-19}  
			\cellcolor{gray!10}RA-MIQA&\cellcolor{OliveGreen!20}\makecell{0.6573 \\ \textcolor{red!100}{\textbf{+13.56\%}}}&\cellcolor{OliveGreen!20}\makecell{0.5823 \\ \textcolor{red!100}{\textbf{+12.05\%}}}&\cellcolor{OliveGreen!20}\makecell{0.2917 \\ \textcolor{red!100}{\textbf{+4.86\%}}}&\cellcolor{OliveGreen!20} \makecell{0.7707 \\ \textcolor{red!100}{\textbf{+13.37\%}}}&\cellcolor{OliveGreen!20}	\makecell{0.7866 \\ \textcolor{red!100}{\textbf{+9.42\%}}}&\cellcolor{OliveGreen!20}\makecell{0.1732 \\ \textcolor{red!100}{\textbf{+11.18\%}}}&\cellcolor{blue!20}\makecell{0.7448 \\ \textcolor{red!100}{\textbf{+4.59\%}}}&\cellcolor{blue!20}\makecell{0.7370 \\ \textcolor{red!100}{\textbf{+4.51\%}}}&\cellcolor{blue!20}\makecell{0.1915 \\ \textcolor{red!100}{\textbf{+4.63\%}}}&\cellcolor{blue!20}\makecell{0.8526 \\ \textcolor{red!100}{\textbf{+0.79\%}}}&\cellcolor{blue!20}\makecell{0.8692 \\ \textcolor{red!100}{\textbf{+0.84\%}}}&\cellcolor{blue!20}\makecell{0.1527 \\ \textcolor{red!100}{\textbf{+2.49\%}}}&\cellcolor{orange!20}\makecell{0.7363 \\ \textcolor{red!100}{\textbf{+2.72\%}}}&\cellcolor{orange!20}\makecell{0.7327 \\ \textcolor{red!100}{\textbf{+2.53\%}}}&\cellcolor{orange!20}\makecell{0.1834 \\ \textcolor{red!100}{\textbf{+2.71\%}}}&\cellcolor{orange!20}\makecell{0.8632 \\ \textcolor{red!100}{\textbf{+1.71\%}}}&\cellcolor{orange!20}\makecell{0.8756 \\ \textcolor{red!100}{\textbf{+1.62\%}}}&\cellcolor{orange!20}\makecell{0.1464 \\ \textcolor{red!100}{\textbf{+4.87\%}}} \stepcounter{rowno} \\  
			\bottomrule 
		\end{tabular}
	} \vspace{-0.7em}
\end{table*}
\subsection{Implementation Details}
The experiments were conducted using the PyTorch framework on NVIDIA GeForce RTX 3090 GPUs. Models were initialized with ImageNet pre-trained weights and trained for 5 epochs with a batch size of 256. The AdamW optimizer was used with an initial learning rate of 1e-4 and weight decay of 1e-4. The learning rate schedule employed a cosine annealing scheduler with a single epoch of warm-up period. Validation was performed twice per epoch. For data preprocessing, the shorter side of each image was resized to 288 pixels. During training, random crops of size 224$\times$224 were extracted, while center cropping to the same size was applied during validation. Data augmentation during training included random horizontal and vertical flips (with a probability of 0.5). The region encoder followed the architecture in~\cite{gao2022luss}, operated on center-cropped inputs of size 288$\times$288, and remained frozen during training. The proposed and baseline MIQA models were optimized using the mean squared error (MSE) loss. 

\subsection{Main Results}
\noindent{\textbf{Evaluation of composite score}}: Table~\ref{tab:resutls} presents the performance of HVS-based and machine-based IQA methods on the MIQD-2.5M. HVS-based FR-IQA methods exhibited limited effectiveness for MIQA tasks. DISTS performed relatively better among FR-IQA metrics but still achieved only modest correlation values~(\eg, SRCC/PLCC of 0.3878/0.3804 for image classification and 0.5266/0.5352 for object detection), while well-used metrics~(\ie, PSNR and SSIM) performed even worse. Similarly, DNN-based NR-IQA approaches~(\ie, HyperIQA and MANIQA) yielded suboptimal results despite being trained on IQA dataset. These results demonstrate the inadequacy of HVS-based IQA metrics in assessing degradation relevant to MVS, underscoring the necessity of efforts for developing MIQA models. For MIQA models evaluated on the million-scale database, EfficientNet variants demonstrated consistent superiority over ResNets, with EfficientNet-b1 outperforming ResNet-18/50 and EfficientNet-b5 achieving further gains with increased capacity. ViT-small exceeded most CNN-based approaches except EfficientNet-b5 in the classification task. The proposed RA-MIQA~(an enhanced ViT-small model) achieved superior performance, improving SRCC/PLCC metrics by 16.76\%/13.44\%, 1.66\%/1.50\%, and 2.76\%/2.47\% over ViT-small in classification, detection, and segmentation tasks, respectively. In addition, all models exhibited lower performance on classification compared to detection and segmentation tasks. This discrepancy arises from the differences in task granularity: detection and segmentation operate at a finer level where models respond more consistently to distortion intensity, whereas classification functioning at a coarser semantic level exhibits non-uniform sensitivity across different degradation magnitudes.  

\begin{table*}[tb]\centering
	\caption{Comparison of HVS-based and machine-based IQA methods across degradation regions, including UD, ROI-DD, and BG-DD. The best results within each category are highlighted with a colored background. The last row reports the performance gains relative to ViT-small.} \vspace{-1.em}
	\label{tab:region_res}
	\setcounter{rowno}{1}
	\resizebox{0.995\textwidth}{!}{
		%\large
		\begin{tabular}{c|CC|CC|CC|CC|CC|CC|CC|CC|CC}
			\toprule 
			\multirow{3}{*}{\textbf{Method}}& \multicolumn{6}{c|}{\textbf{Image Classification}}&\multicolumn{6}{c|}{\textbf{Object Detection}} &\multicolumn{6}{c}{\textbf{Instance Segmentation}}  \\   \cmidrule{2-19} 
			&\multicolumn{2}{c|}{\textbf{UD}}  &\multicolumn{2}{c|}{\textbf{ROI-DD}} &\multicolumn{2}{c|}{\textbf{BG-DD}}  &\multicolumn{2}{c|}{\textbf{UD}}  &\multicolumn{2}{c|}{\textbf{ROI-DD}} &\multicolumn{2}{c|}{\textbf{BG-DD}}&\multicolumn{2}{c|}{\textbf{UD}}  &\multicolumn{2}{c|}{\textbf{ROI-DD}} &\multicolumn{2}{c}{\textbf{BG-DD}} \\  
			\tiny &\textbf{SRCC}$\uparrow$ & \textbf{PLCC}$\uparrow$ &\textbf{SRCC}$\uparrow$ & \textbf{PLCC}$\uparrow$ &\textbf{SRCC}$\uparrow$ & \textbf{PLCC}$\uparrow$ &\textbf{SRCC}$\uparrow$ & \textbf{PLCC}$\uparrow$ &\textbf{SRCC}$\uparrow$ & \textbf{PLCC}$\uparrow$ &\textbf{SRCC}$\uparrow$ & \textbf{PLCC}$\uparrow$ &\textbf{SRCC}$\uparrow$ & \textbf{PLCC}$\uparrow$ &\textbf{SRCC}$\uparrow$ & \textbf{PLCC}$\uparrow$ &\textbf{SRCC}$\uparrow$ & \textbf{PLCC}$\uparrow$ \stepcounter{rowno} \\   \midrule
			PSNR&0.3761 &  0.3807&0.2977 &0.2771&0.1322&0.1253 &0.5172& 0.5325 &0.3070& 0.3415  &0.2277&0.2490& 0.5266&0.5414&0.3021&0.3343&0.2431& 0.2697  \stepcounter{rowno}\\
			SSIM~\cite{ssim}&0.4588 &  0.4590&0.3893&0.3596 &0.1672&0.1679&0.6228 & 0.6312 &0.4756& 0.4831  &0.3309&0.3505& 0.6250&0.6334&0.4592&0.4653&0.3451& 0.3646  \stepcounter{rowno}\\
			VSI~\cite{vsi}&0.4854 &  0.5029&0.4217&0.3920 &0.1839&0.1847&0.6452 & 0.6585 &0.4831& 0.4900 &0.3573&0.3788& 0.6502&0.6622&0.4741&0.4806&0.3764& 0.3958 \stepcounter{rowno} \\ 
			LPIPS~\cite{lpips}&0.5061 &  0.5434&0.4114&0.3835 &0.1903&0.2035&\cellcolor{blue!20}0.7255 & \cellcolor{blue!20}0.7562 &\cellcolor{blue!20}0.6128&\cellcolor{blue!20} 0.6099  &\cellcolor{blue!20}0.4028&\cellcolor{blue!20}0.4305&\cellcolor{orange!20} 0.7355&\cellcolor{orange!20}0.7638&\cellcolor{orange!20}0.6005&\cellcolor{orange!20}0.5974&\cellcolor{orange!20}0.4284& \cellcolor{orange!20}0.4527 \stepcounter{rowno} \\
			DISTS~\cite{dists}&\cellcolor{OliveGreen!20}0.5122 & \cellcolor{OliveGreen!20}0.5391&\cellcolor{OliveGreen!20}0.4429&\cellcolor{OliveGreen!20}0.4001 &\cellcolor{OliveGreen!20}0.2210&\cellcolor{OliveGreen!20}0.2310& 0.6706&0.6922&0.5314&0.5301&0.3890& 0.4210& 0.6792&0.6986&0.5283&0.5276&0.4152& 0.4432 \stepcounter{rowno}\\ \cdashline{1-19} \addlinespace[3pt]
			HyperIQA~\cite{hyperiqa}&0.3174 &  0.3049&0.2111&0.2019 &0.1600&0.1767&0.5290 & 0.5401 &0.3951& 0.4032  &0.3694&0.4118& 0.5343&0.5452&0.3851&0.3941&0.3766& 0.4189  \stepcounter{rowno}\\
			MANIQA~\cite{maniqa}&0.4128 &  0.3967&0.3144&0.2935 &0.2338&0.2485&0.5368 & 0.5471 &0.3970& 0.4017  &0.4069&0.4392& 0.5450&0.5538&0.3954&0.3999&0.4219& 0.4538  \stepcounter{rowno}\\ 
			\midrule 
			ResNet-18~\cite{resnet}&0.5831 &0.6538  &0.5419& 0.5328& 0.3639& 0.3599&0.8115& 0.8482& 0.7736& 0.7680& 0.6361& 0.6796&0.8159 &0.8535& 0.7755&0.7716&0.6439 &0.6877 \stepcounter{rowno}\\
			ResNet-50~\cite{resnet}&0.6188&0.6780&0.5831& 0.5704& 0.4263& 0.4209&0.8267& 0.8599& 0.7963& 0.7914& 0.6609& 0.7023 &0.8290&0.8640 &0.7951&0.7912&0.6574&0.7001  \stepcounter{rowno}\\
			EfficientNet-b1~\cite{efficientnet}	&0.6382&0.7022&0.6165&0.6038&0.4744	&0.4749&0.8253	&0.8597&0.7978& 0.7929&0.6693&0.7093&0.8297	&0.8652&0.7991&0.7955&0.6780&0.7169  \stepcounter{rowno}\\
			EfficientNet-b5~\cite{efficientnet}	&0.6770&0.7261&0.6526&0.6348&0.5319	&0.5189&0.8374	&0.8684&0.8096&0.8050&0.6795& 0.7156&0.8443&0.8744&0.8067& 0.8030&0.6872&0.7234 \stepcounter{rowno} \\
			ViT-small~\cite{vit}&0.6522 &0.7096  &0.6246& 0.6093& 0.4813& 0.4682&0.8489& 0.8771& 0.8169& 0.8117& 0.7015& 0.7344&0.8507&0.8797 &0.8130&0.8094&0.6974&0.7325  \stepcounter{rowno}\\ 
			\cmidrule{1-19} 
			\cellcolor{gray!10}	RA-MIQA&\cellcolor{OliveGreen!20}\makecell{0.7363 \\ \textcolor{red!100}{\textbf{+12.89\%}}}&\cellcolor{OliveGreen!20} \makecell{0.7653 \\ \textcolor{red!100}{\textbf{+7.85\%}}}&\cellcolor{OliveGreen!20}\makecell{0.7219 \\ \textcolor{red!100}{\textbf{+15.58\%}}}&\cellcolor{OliveGreen!20}\makecell{0.7016 \\ \textcolor{red!100}{\textbf{+15.15\%}}}&\cellcolor{OliveGreen!20}\makecell{0.6152 \\ \textcolor{red!100}{\textbf{+27.82\%}}}&\cellcolor{OliveGreen!20}\makecell{0.5801 \\ \textcolor{red!100}{\textbf{+23.90\%}}}&\cellcolor{blue!20}\makecell{0.8590 \\ \textcolor{red!100}{\textbf{+1.19\%}}}&\cellcolor{blue!20}\makecell{0.8849 \\ \textcolor{red!100}{\textbf{+0.89\%}}}&\cellcolor{blue!20}\makecell{0.8316 \\ \textcolor{red!100}{\textbf{+1.80\%}}}&\cellcolor{blue!20}\makecell{0.8267 \\ \textcolor{red!100}{\textbf{+1.85\%}}}&\cellcolor{blue!20}\makecell{0.7192 \\ \textcolor{red!100}{\textbf{+2.52\%}}}&\cellcolor{blue!20}\makecell{0.7490 \\ \textcolor{red!100}{\textbf{+1.99\%}}}&\cellcolor{orange!20}\makecell{0.8665 \\ \textcolor{red!100}{\textbf{+1.86\%}}}&\cellcolor{orange!20}\makecell{0.8927 \\ \textcolor{red!100}{\textbf{+1.48\%}}}&\cellcolor{orange!20}\makecell{0.8341 \\ \textcolor{red!100}{\textbf{+2.60\%}}}&\cellcolor{orange!20}\makecell{0.8305 \\ \textcolor{red!100}{\textbf{+2.61\%}}}&\cellcolor{orange!20}\makecell{0.7301 \\ \textcolor{red!100}{\textbf{+4.69\%}}}&\cellcolor{orange!20}\makecell{0.7615 \\ \textcolor{red!100}{\textbf{+3.96\%}}} \stepcounter{rowno}\\ 
			\bottomrule  
		\end{tabular}
	} \vspace{-.5em}
\end{table*}
\begin{table*}[t]
	\centering
	\caption{Comparison of MIQA methods across different distortion types. The two best-performing distortions are highlighted with a \textcolor{OliveGreen!100}{green} background, while the two worst-performing are marked with a \textcolor{red!100}{red} background.} \vspace{-1.em}
	\label{tab:dist_type}
	\setcounter{rowno}{1}
	\resizebox{0.995\textwidth}{!}{
		\Large
		\begin{tabular}{c|c|cc|cc|cc|cc|cc|cc|cc|cc|cc|cc}
			\toprule 
			\multirow{2}{*}{\textbf{Task}}&\multirow{2}{*}{\textbf{Method}}& \multicolumn{2}{c|}{\textbf{Contrast}}& \multicolumn{2}{c|}{\textbf{Darkness}}& \multicolumn{2}{c|}{\textbf{Pixelate}}&\multicolumn{2}{c|}{\textbf{JPEG}}&\multicolumn{2}{c|}{\textbf{Motion Blur}} &\multicolumn{2}{c|}{\textbf{Defocus Blur}} &\multicolumn{2}{c|}{\textbf{Glass Blur}} &\multicolumn{2}{c|}{\textbf{Fog}} &\multicolumn{2}{c|}{\textbf{Snow}}&\multicolumn{2}{c}{\textbf{Gaussian Noise}} \\   
			&	\tiny &\textbf{SRCC}$\uparrow$ & \textbf{PLCC}$\uparrow$&\textbf{SRCC}$\uparrow$ & \textbf{PLCC}$\uparrow$ &\textbf{SRCC}$\uparrow$ & \textbf{PLCC}$\uparrow$ &\textbf{SRCC}$\uparrow$ & \textbf{PLCC}$\uparrow$ &\textbf{SRCC}$\uparrow$ & \textbf{PLCC}$\uparrow$ &\textbf{SRCC}$\uparrow$ & \textbf{PLCC}$\uparrow$ &\textbf{SRCC}$\uparrow$ & \textbf{PLCC}$\uparrow$ &\textbf{SRCC}$\uparrow$ & \textbf{PLCC}$\uparrow$ &\textbf{SRCC}$\uparrow$ & \textbf{PLCC}$\uparrow$ &\textbf{SRCC}$\uparrow$ & \textbf{PLCC}$\uparrow$  \stepcounter{rowno} \\   \midrule 
			%\multicolumn{21}{l}{{\textbf{Performance comparison of various methods for MIQA in image classification}.}}  \\ 
			%\midrule
			%	\cdashline{1-21} \addlinespace[3pt]
			\multirow{7}{*}{\rotatebox{90}{Image~Classification}}&ResNet-18~\cite{resnet}	&0.4048	&0.3646 &0.4016&0.4275&0.6320 & 0.6754& 0.4090 & 0.4093& 0.5407&0.5470&0.4632 & 0.4556	& 0.5012& 0.5237& 0.4450 & 0.4647&0.5247 &0.4980&0.6181& 0.6449\\
			&ResNet-50~\cite{resnet}	&0.4666 & 0.4191 &0.4585 & 0.4700&0.6653 & 0.7003&0.4632  &0.4492&0.5912 & 0.5954&0.5112 & 0.4985&0.5461&  0.5632&0.4971 & 0.5074&0.5561 & 0.5292&0.6548 & 0.6724 \stepcounter{rowno}\\
			&	EfficientNet-b1~\cite{efficientnet}	&0.4925 &0.4632&0.4881&  0.5117&0.6847 & 0.7202&0.5088&  0.4986&0.6206 & 0.6252 & 0.5577 & 0.5419&0.5850& 0.5950&0.5221 & 0.5381&0.6202 & 0.5967&0.6769 & 0.6909\\% \cmidrule{2-14}
			&EfficientNet-b5~\cite{efficientnet}	&0.5539 & 0.5124&0.5512 & 0.5630&0.7200&  0.7444&0.5618 & 0.5455& 0.6622 & 0.6583&0.5938&0.5728&0.6234&0.6238&0.5740& 0.5835&0.6393&0.6140&0.7133& 0.7231 \stepcounter{rowno}\\ 
			&	ViT-small~\cite{vit}	&0.5150&  0.4678&0.5002 & 0.5134&0.6964& 0.7267&0.5145& 0.5036& 0.6274& 0.6283&0.5584& 0.5476&0.5865& 0.6031&0.5455& 0.5502&0.6072& 0.5777&0.6927& 0.7061 \stepcounter{rowno}\\  
			&RA-MIQA &0.6242& 0.5743&0.6323&0.6191&0.7695&0.7790&0.6398 & 0.6019&0.7270 & 0.7156&0.6813 & 0.6563&0.7006 & 0.6940&0.6462 & 0.6319&0.7118 & 0.6801&0.7641 & 0.7671 \stepcounter{rowno} \\ \cmidrule{2-22} 
			&	\textbf{Average}&\cellcolor{red!20}0.5095&\cellcolor{red!30}0.4669&\cellcolor{red!30}0.5053&	0.5175&\cellcolor{OliveGreen!30}0.6947&\cellcolor{OliveGreen!30}0.7243&	0.5162&\cellcolor{red!20}0.5014&	0.6282&	0.6283&	0.5609&	0.5455&	0.5905&	0.6005&	0.5383&	0.5460&	0.6099&	0.5826&\cellcolor{OliveGreen!20}0.6867&\cellcolor{OliveGreen!20}0.7008   \\  
			\toprule  
			\multirow{7}{*}{\rotatebox{90}{Object~Detection}}&ResNet-18~\cite{resnet}	&0.6361& 0.6526&0.6452& 0.6864&0.8572& 0.8402&0.7477 & 0.7818&0.7732& 0.7666&0.6915& 0.7035&0.7106& 0.7305&0.5520&  0.5757&0.6961&0.6915&0.8399& 0.8252 \stepcounter{rowno}\\
			&ResNet-50~\cite{resnet}	&0.6654& 0.6842&0.6739& 0.7157&0.8719& 0.8553&0.7669& 0.8017&0.7877 & 0.7827&0.7128 & 0.7275&0.7346 & 0.7570&0.5904 & 0.6198&0.7216  &0.7197&0.8518 & 0.8379	\stepcounter{rowno}\\
			&	EfficientNet-b1~\cite{efficientnet}	&0.6706&  0.6889&0.6767&  0.7186&0.8766&0.8576&0.7695&  0.8046& 0.7945& 0.7898&0.7188 & 0.7336&0.7375&  0.7591&0.5841 & 0.6131&0.7332&  0.7304&0.8521& 0.8395 \stepcounter{rowno}\\% \cmidrule{2-14}
			&EfficientNet-b5~\cite{efficientnet}	&0.6873& 0.7069&0.6894 & 0.7283&0.8813&  0.8629&0.7840 & 0.8159& 0.8041&  0.7999&0.7308 & 0.7464& 0.7487 & 0.7712&0.6099 & 0.6359&0.7335 & 0.7330& 0.8576& 0.8440	\stepcounter{rowno}\\ 
			&	ViT-small~\cite{vit}	&0.7075 & 0.7235 &0.7078 & 0.7445	&0.8854 & 0.8663&0.7994 & 0.8263& 0.8148 &0.8093&0.7398 & 0.7529 &0.7597 & 0.7791&0.6381 & 0.6605& 0.7571&  0.7544&0.8699 & 0.8565 \stepcounter{rowno} \\  
			&RA-MIQA &0.7239 & 0.7377&0.7283&0.7633&0.8941&0.8763&0.8076 & 0.8323&0.8305 & 0.8256&0.7603 & 0.7738&0.7776 & 0.7959&0.6582 & 0.6771&0.7750 & 0.7720&0.8784 & 0.8656 \stepcounter{rowno} \\ \cmidrule{2-22} 
			&	\textbf{Average}&\cellcolor{red!20}0.6818&\cellcolor{red!20}0.6990&0.6869&0.7261&\cellcolor{OliveGreen!30}0.8778&\cellcolor{OliveGreen!30}0.8598&	0.7792&	0.8104&	0.8008&	0.7957&	0.7257&	0.7396&	0.7448&	0.7655&\cellcolor{red!30}0.6055&\cellcolor{red!30}0.6304&	0.7361&	0.7335&\cellcolor{OliveGreen!20}0.8583&\cellcolor{OliveGreen!20}0.8448\\  
			\toprule  
			\multirow{7}{*}{\rotatebox{90}{Instance~Segmentation}}&	ResNet-18~\cite{resnet} &0.6452&  0.6673&0.6608 & 0.7031&0.8653 & 0.8499&0.7504 & 0.7885&0.7767 & 0.7708&0.7023  &0.7145&0.7188 & 0.7370&0.5586 & 0.5866&0.6797 & 0.6814&0.8382 &  0.8246 \stepcounter{rowno}	\\
			&	ResNet-50~\cite{resnet}&0.6622&  0.6868&0.6776 & 0.7197&0.8770&  0.8589&0.7646 & 0.8025&0.7964&  0.7910&0.7185 & 0.7321&0.7362 & 0.7568&0.5894 & 0.6180&0.6997 & 0.7009 &0.8465 & 0.8325	\stepcounter{rowno}\\
			&EfficientNet-b1~\cite{efficientnet}	&0.6779 & 0.6992&0.6950&  0.7325&0.8823&  0.8637&0.7727&  0.8097&0.8034 & 0.7993&0.7283 & 0.7416&0.7439&  0.7646&0.6002&  0.6295&0.7157&  0.7166&0.8492 & 0.8366 \stepcounter{rowno} \\% \cmidrule{2-14}
			&EfficientNet-b5~\cite{efficientnet}&0.6913&  0.7122& 0.7081 & 0.7419&0.8871&  0.8684&0.7869 & 0.8206&0.8073  &0.8032 & 0.7413 & 0.7540&0.7557 & 0.7754&0.6246 & 0.6442&0.7192 & 0.7213&0.8533 & 0.8408 \stepcounter{rowno} \\ 
			&	ViT-small~\cite{vit}	&0.7068  &0.7263&0.7168 & 0.7531&0.8899 & 0.8721&0.7967 & 0.8276&0.8146&  0.8096&0.7401  &0.7530& 0.7581&  0.7772&0.6395& 0.6641&0.7304 &0.7320&0.8629 & 0.8503 \stepcounter{rowno} \\  
			&RA-MIQA &0.7381 & 0.7561&0.7467&0.7778&0.9029&0.8859&0.8141 & 0.8411&0.8376 & 0.8332&0.7695 & 0.7826&0.7851 & 0.8026&0.6824 & 0.7029&0.7663 & 0.7670&0.8755 & 0.8636 \stepcounter{rowno} \\ \cmidrule{2-22}    
			&\textbf{Average}&\cellcolor{red!20}0.6869&\cellcolor{red!20}0.7080&0.7008&0.7380&\cellcolor{OliveGreen!30}0.8841&\cellcolor{OliveGreen!30}0.8665&0.7809&	0.8150&	0.8060&	0.8012&	0.7333&	0.7463&	0.7496&	0.7689&	\cellcolor{red!30}0.6158&\cellcolor{red!30}0.6409&	0.7185&	0.7199&\cellcolor{OliveGreen!20}0.8543&\cellcolor{OliveGreen!20}0.8414 \\\addlinespace[3.5pt]
			\bottomrule
		\end{tabular}
	} \vspace{-.5em}
\end{table*}
\noindent\textbf{Evaluation of consistency and accuracy scores}:
Table~\ref{tab:cons_acc} reports the independent performance of retrained representative models (\eg, ResNet50, EfficientNet-b5, and ViT-small) on consistency and accuracy metrics. Among HVS-based IQA methods, DISTS achieves optimal or near-optimal results in both metrics; however, its performance remains insufficient for effectively quantifying machine vision. The retrained models demonstrate superiority over HVS-based methods and underscore the necessity of adopting machine-specific IQA paradigms. Our RA-MIQA model further elevates performance across all tasks and evaluation metrics, improving SRCC over ViT-small by 13.56\%/13.37\%, 4.59\%/0.79\%, and 2.72\%/1.71\% on accuracy/consistency scores for classification, detection, and segmentation tasks, respectively. These results indicate that the MIQA models exhibit higher precision in estimating consistency scores compared to accuracy scores. This disparity reflects their fundamental difference: consistency scores, focusing exclusively on prediction changes relative to original predictions (regardless of their correctness), exhibit a more linear relationship with image quality. In contrast, accuracy scores capture both degradation effects and model limitations (\eg, misclassifications on pristine images), leading to weaker correlations and higher errors. Despite these differences, both metrics provide vital complementary insights: consistency evaluates a model's stability under perturbations, while accuracy measures its practical ability to maintain correct predictions amid interference.

\noindent\textbf{Evaluation of different degradation regions}:
Table~\ref{tab:region_res} shows that MIQA methods consistently perform best with UD compared to ROI-DD and BG-DD, with BG-DD presenting the greatest challenge. Using ViT-small as a representative example, classification task SRCC/PLCC metrics under UD exceed ROI-DD by 4.42\%/16.46\% and BG-DD by 35.51\%/51.56\%. Similarly, for object detection, UD metrics surpass ROI-DD by 3.92\%/8.06\% and BG-DD by 21.01\%/19.43\%, while in instance segmentation, UD outperforms ROI-DD by 4.64\%/8.69\% and BG-DD by 21.98\%/20.10\%. Our proposed RA-MIQA, with its enhanced distortion region awareness, demonstrates superior assessment capabilities for ROI-DD and BG-DD, with performance gains more pronounced in these challenging regions compared to UD. These performance variations across UD, ROI-DD, and BG-DD stem from the interplay between distortions, critical features, and contextual information. UD exhibits consistent spatial patterns, leading to a stronger correlation between degradation and quality scores. In contrast, ROI-DD and BG-DD introduce spatially heterogeneous distortions, increasing perceptual complexity for vision models. ROI-DD directly affects critical features, yet the preserved background information provides compensatory contextual cues, leading to a modest impact on the degradation-quality correlation. Conversely, for BG-DD, compromised background undermines the model’s contextual reasoning capabilities, causing performance degradation through indirect mechanisms. This indirect relationship generates more complex degradation-quality correlations compared to both UD and ROI-DD, increasing the difficulty of the MIQA task. 
 
\begin{figure}[t] 
	\centering
	\includegraphics[width=0.495\textwidth]{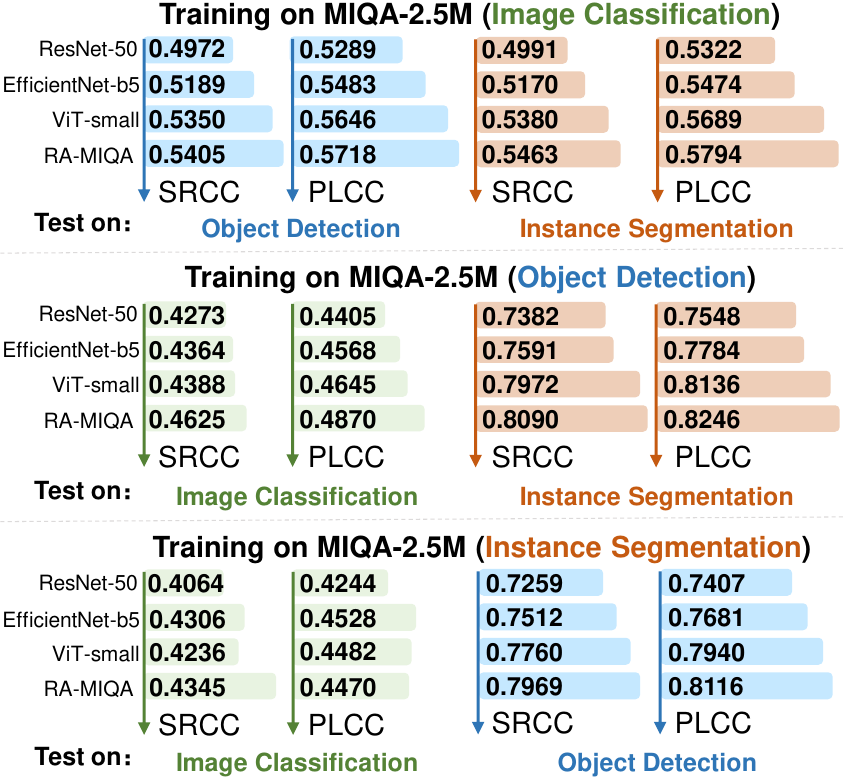} 
	\caption{Cross-task validation results of MIQA models.} \vspace{-1.em}
	\label{fig:cross_database}
\end{figure}

\noindent\textbf{Evaluation of degradation in different types}:~Table~\ref{tab:dist_type} shows the performance of the MIQA methods in a variety of degradation types, revealing their ability to be evaluated in a variety of distortion scenarios. MIQA methods demonstrate superior performance for Pixelation and Gaussian Noise across all vision tasks, indicating strong correlations between these distortions and quality scores that facilitate effective model training and evaluation. Conversely, MIQA methods struggle significantly with contrast distortions, while fog presents particular challenges for object detection and instance segmentation tasks. Darkness and JPEG compression similarly cause substantial performance degradation in image classification. The observed performance variations reflect the differences in how distortion types influence machine perception. Pixel-level degradations (\eg, Pixelation, Gaussian Noise) produce direct and predictable effects on feature extraction processes, resulting in stronger correlations with MIQA scores. In contrast, global distortions (\eg, contrast adjustments) or those introducing complex artifacts (\eg, JPEG compression) affect visual models through more sophisticated pathways, creating intricate degradation-performance relationships that current MIQA approaches struggle to accurately model. Identifying and mitigating these challenging degradation types demands particular attention to prevent potentially severe machine vision failures in critical real-world applications.
 
\begin{figure}[t] \centering
	\includegraphics[width=0.48\textwidth]{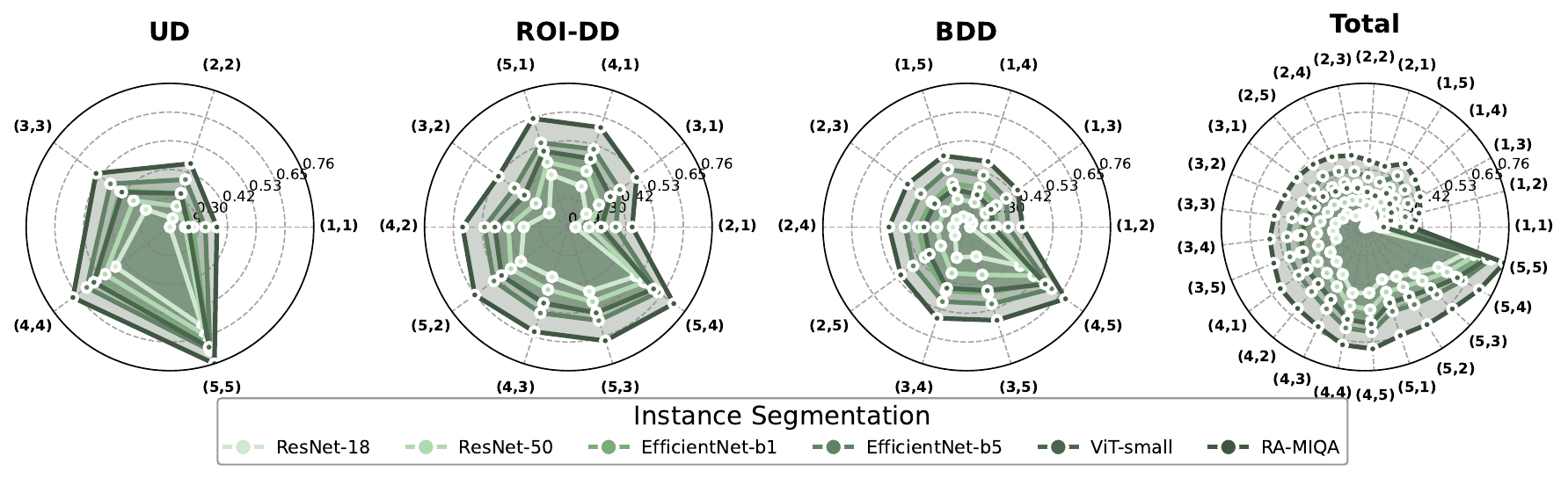}
	\includegraphics[width=0.48\textwidth]{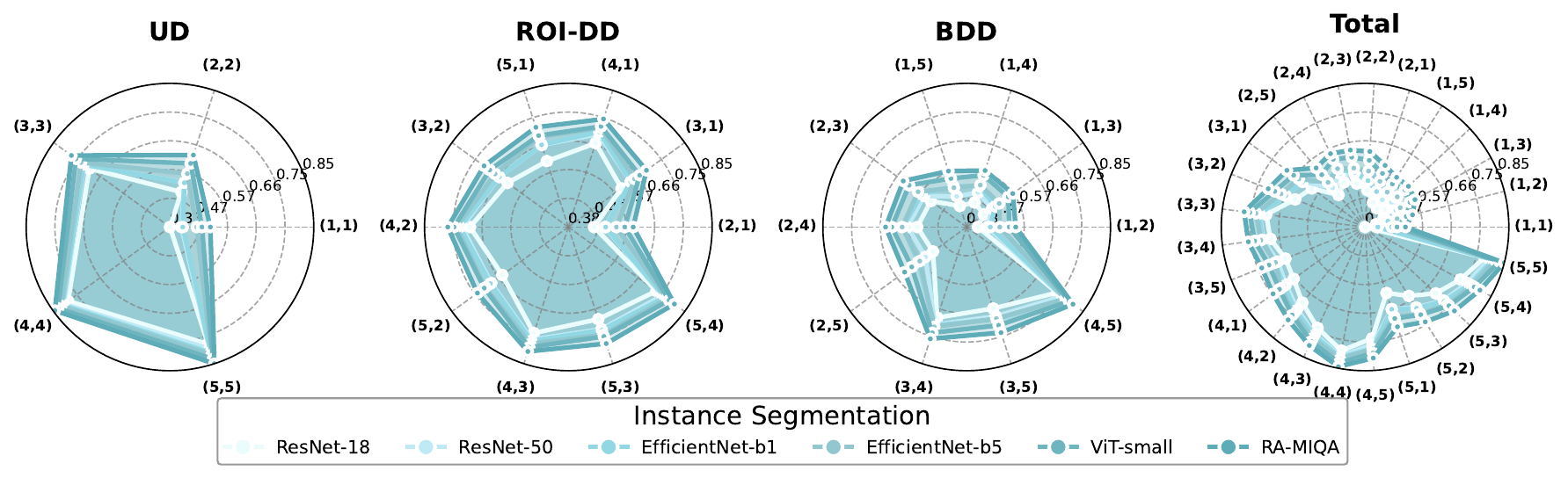}  
	\includegraphics[width=0.48\textwidth]{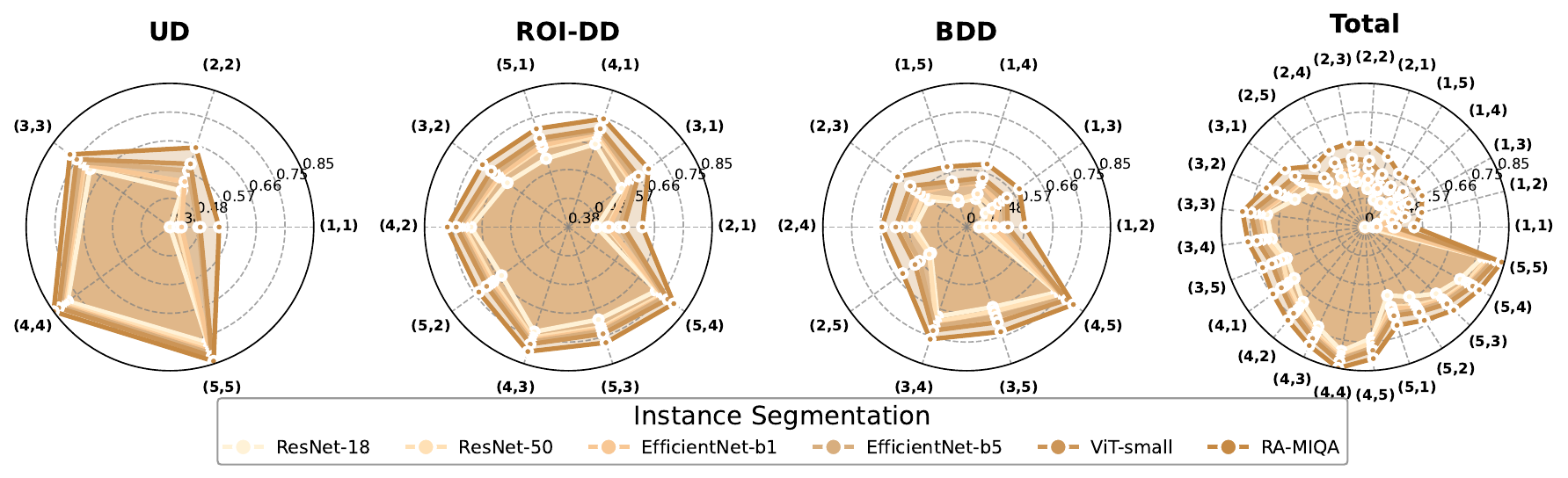}
	\caption{The PLCC values of MIQA models across degradation regions and levels.} \vspace{-1em}
	\label{fig:region_radar}  
\end{figure}
\subsection{Discussion}\label{sec:discussion}
 
\noindent \textbf{Cross-task generalization}:~Figure~\ref{fig:cross_database} reveals that MIQA models exhibit notable task-dependent generalization behaviors. For instance, RA-MIQA trained on classification achieves only moderate SRCC/PLCC scores on detection (0.5405/0.5718) and segmentation (0.5463/0.5794). Conversely, training on detection yields strong performance on segmentation (0.8090/0.8246), but results in poor generalization back to classification (0.4625/0.4870). Similar trends are observed for models trained on segmentation. These results reflect the differences in task-specific quality perception. Fine-grained tasks rely on precise spatial details and local feature integrity, creating aligned sensitivities to specific distortions that facilitate effective cross-generalization between detection and segmentation. In contrast, coarse-grained tasks prioritize global semantic features with different distortion sensitivities, limiting their transferability to fine-grained contexts. This task-specific quality perception challenges the feasibility of universal MIQA models across tasks of varying granularity. In contrast, the strong generalization observed between detection and segmentation~(both fine-grained tasks) shows the potential for cross-task adaptation, enabling MIQA model reuse with minimal retraining. 

\noindent\textbf{Performance on degradation regions and levels}:~Figure~\ref{fig:region_radar} illustrates the PLCC values of MIQA models across varying distortion regions and intensity levels. For UD, MIQA models demonstrate improved performance with increasing distortion severity across the three tasks. This phenomenon emerges because severe distortions more consistently impact machine vision performance, creating clearer degradation-quality relationships. Conversely, slight distortions introduce greater uncertainty that machines may respond inconsistently to minor visual perturbations, substantially increasing the difficulty of MIQA evaluation. Similar trends appear in both ROI-DD and BG-DD, where models perform reliably under severe conditions but struggle with slight degradations, as confirmed by the aggregated ``Total" results. Despite these challenges, the proposed RA-MIQA enhances distortion sensitivity under these conditions. Moreover, ROI-DD is easier to evaluate than BG-DD because it directly affects the quality of critical regions, enabling straightforward detection of quality degradation. In contrast, background distortion influences machine vision more subtly. For instance, as demonstrated in the “Total” subfigure, when the ROI distortion level remains constant, variations in the background distortion level yield divergent results. In other words, the background distortion has an impact on machine perception making the evaluation of the MIQA model different. This indirect effect via background disruption undermines contextual inference, posing a greater challenge than assessing direct distortions in key regions.

\begin{figure}[t!] \centering  
	\includegraphics[width=0.495\textwidth]{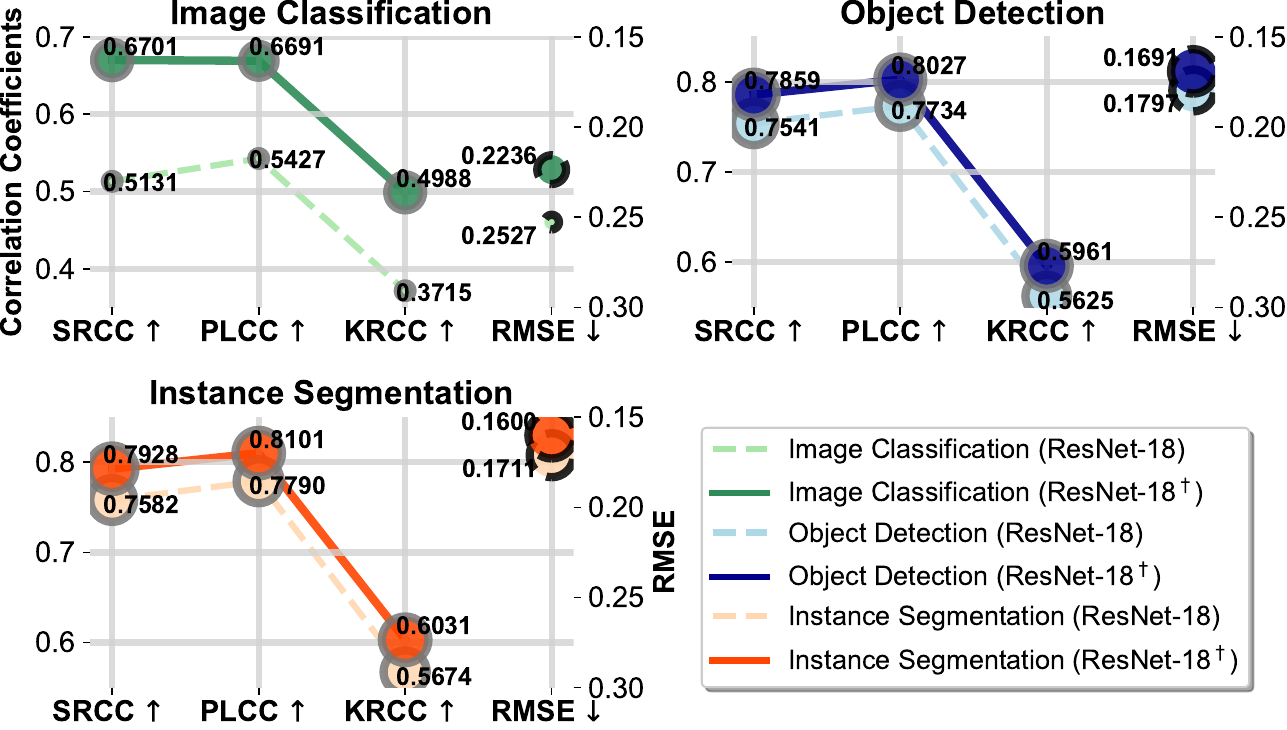}  
	\caption{Ablation study of the region encoder. Performance improvements across all evaluation metrics for three MIQA tasks using ResNet-18 as a baseline. Correlations (left axis) and RMSE (right axis) are shown in the dual-axis plot.} \vspace{-1.em}
	\label{fig:resnet18_plus}
\end{figure}
\noindent\textbf{Validity of region encoder}: To further validate the effectiveness of the region encoder in extracting degradation region-aware features, we conducted experiments using ResNet-18 as the baseline model. The features output by the region encoder were concatenated with the baseline model's output features for both training and prediction purposes. As shown in Figure~\ref{fig:resnet18_plus}, the enhanced model (ResNet-18$^\dag$) significantly outperformed the baseline across all MIQA tasks, yielding SRCC/PLCC improvements of 30.61\%/23.29\% for classification, 4.22\%/3.79\% for detection, and 4.56\%/3.99\% for segmentation. By incorporating region-specific distortion features spanning both ROI and background areas, the model achieves more precise quality assessment by effectively capturing spatially varying distortions in MVS.

\begin{figure*}[thbp] 
	\centering 
	\includegraphics[width=0.99\textwidth]{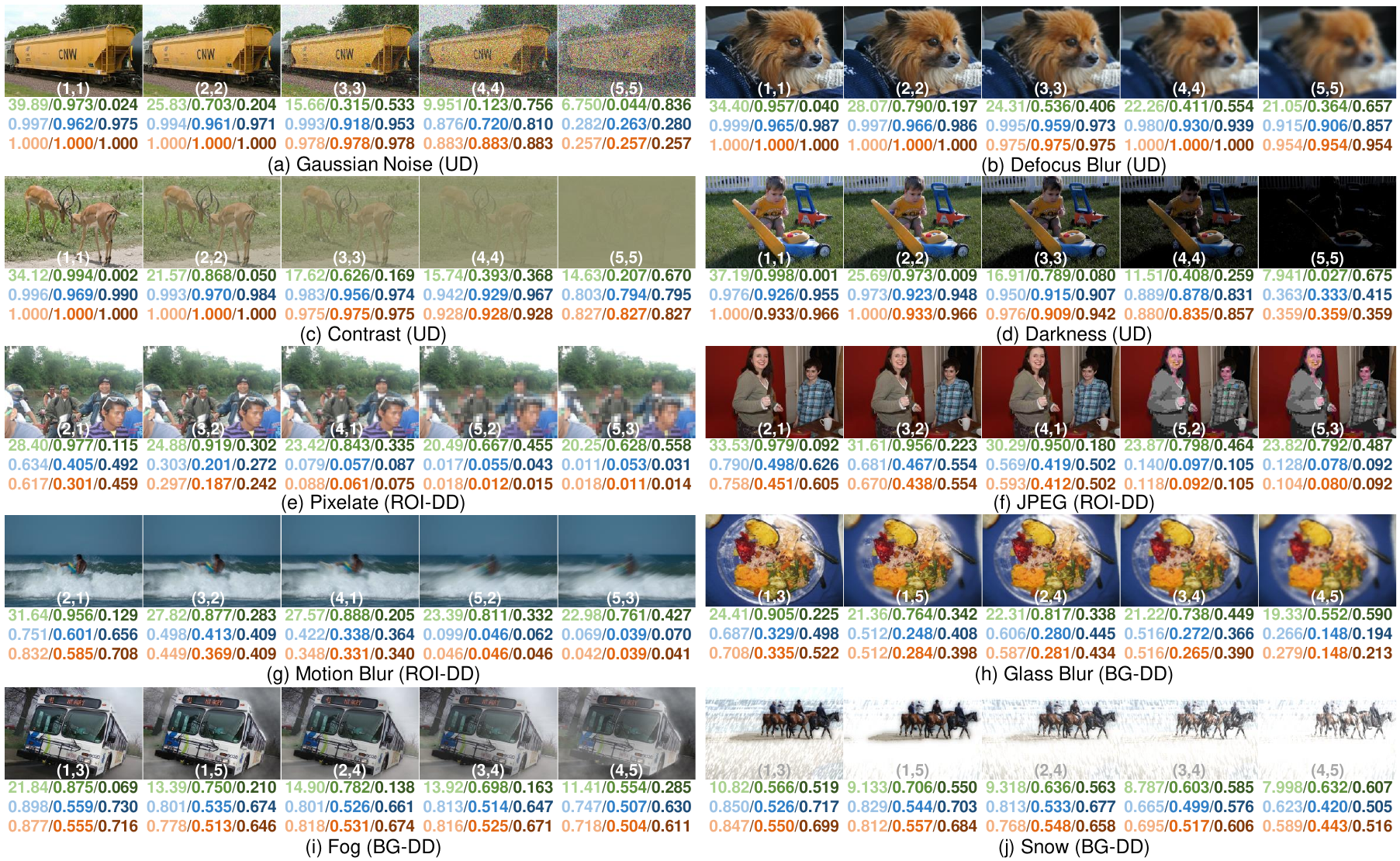} %\vspace{-1.em}
	\caption{Sample illustration showcasing the corresponding \textcolor{OliveGreen!40}{\textbf{PSNR}}, \textcolor{OliveGreen!90}{\textbf{SSIM}}, and \textcolor{DeepOliveGreen}{\textbf{LPIPS}} values, along with the \textbf{predicted} \textcolor{NavyBlue!40}{\textbf{Consistency}}, \textcolor{NavyBlue!90}{\textbf{Accuracy}}, and \textcolor{MidnightBlue}{\textbf{Composite}} scores, and their respective \textbf{ground-truth} \textcolor{Bittersweet!40}{\textbf{Consistency}}, \textcolor{Bittersweet!90}{\textbf{Accuracy}}, and \textcolor{Brown}{\textbf{Composite}} scores. Panels (a)–(d), (e)–(g), and (h)–(j) present example images along with MIQA-related scores for image classification under UD, object detection under ROI-DD, and instance segmentation under BG-DD, respectively. Numbers in parentheses indicate distortion severity in ROI and background regions. Note that lower LPIPS indicates higher perceptual quality, whereas higher values are preferred for other metrics.} \vspace{-1.em}
	\label{fig:dist_examples}
\end{figure*}
\noindent \textbf{Examples}: Figure~\ref{fig:dist_examples} showcases some samples from the MIQD-2.5M database, illustrating diverse degradation types, regions, and severity levels. Each example includes scores from HVS-based IQA methods (PSNR, SSIM, and LPIPS), our proposed RA-MIQA, and ground-truth labels. The results highlight the limitations of HVS-based methods in reflecting MVS performance. In example (a) at distortion level (3,3), SSIM reports a low score (0.315), while the ground-truth consistency and accuracy are high (0.978), accurately captured by RA-MIQA (0.993/0.918). This indicates that the MVS remains robust despite perceptual degradation, which HVS-based metrics fail to capture. Conversely, in example (g) at level (5,1), SSIM gives a high score (0.811), yet model performance is poor (0.046), again well estimated by RA-MIQA (0.099/0.046). Here, distortions in the ROI significantly impair recognition, while the background remains visually clean—misleading HVS-based metrics. These cases underscore the need for machine-centric IQA approaches that account for task-relevant distortions beyond human perceptual quality.

\section{Conclusion}\label{sec:conc}
This study introduces a paradigm for machine image quality assessment (MIQA). To understand machine vision quality perception and develop effective MIQA models, we established the MIQD-2.5M, comprising 2.5 million samples across 75 vision models, 250 degradation scenarios, and three representative vision tasks, with each sample annotated with consistency and accuracy metrics. We designed a degradation region-aware model~(RA-MIQA) that demonstrates superior performance across multiple dimensions of evaluation. Our investigation reveals several key findings: HVS-based IQA metrics fundamentally fail to capture machine perception requirements, as human visual sensitivity differs markedly from machine perception mechanisms, thereby highlighting the critical need for MIQA approaches. Fine-grained tasks with pixel- or region-level localization (\eg, detection, segmentation) show strong cross-task generalization but poor transferability to classification, revealing the differences in visual information processing across tasks. Even specialized MIQA models face significant challenges, particularly with background distortions and subtle degradations that produce complex, non-linear effects on machine perception. These gaps create opportunities for developing more effective MIQA models. We hope this work establishes essential foundations for improving machine-oriented image processing methods and contributes to more reliable vision systems spanning critical applications such as autonomous driving, medical diagnosis, and industrial inspection.

\bibliographystyle{spmpsci}
\bibliography{egbib}

\end{document}